\documentclass{article}

\PassOptionsToPackage{numbers, compress}{natbib}

\usepackage[main, final]{neurips_2026}

\usepackage[utf8]{inputenc} 
\usepackage[T1]{fontenc}    
\usepackage{hyperref}       
\usepackage{url}            
\usepackage{booktabs}       
\usepackage{amsfonts}       
\usepackage{nicefrac}       
\usepackage{microtype}      
\usepackage{xcolor}         
\usepackage{amsmath}
\usepackage{amssymb}
\usepackage{mathtools}

\usepackage{multirow} 
\usepackage{amsthm}
\usepackage{enumitem}

\usepackage[normalem]{ulem}
\usepackage{hyperref}
\usepackage{titletoc}
\usepackage{subcaption}
\usepackage{wrapfig}
\usepackage{makecell}
\definecolor{darkblue}{RGB}{0,0,139}

\title{XDecomposer: Learning Prior-Free Set Decomposition for Multiphase X-ray Diffraction}
\author{
	Hanyu Gao$^{1,2,*}$, Bin Cao$^{3,4,*,\dagger}$, Yunyue Su$^{1}$, Tong-Yi Zhang$^{3}$, Qiang Liu$^{1,\dagger}$ \\
	$^{1}$New Laboratory of Pattern Recognition (NLPR), Institute of Automation,\\
	Chinese Academy of Sciences (CASIA) \\
	$^{2}$School of Advanced Interdisciplinary Sciences,\\
	University of Chinese Academy of Sciences \\
    $^{3}$Guangzhou Municipal Key Laboratory of Materials Informatics, \\
    The Hong Kong University of Science and Technology (Guangzhou)\\
    $^{4}$AI Lab, The Yangtze River Delta.
}

\begin{document}
\maketitle

\begingroup
\renewcommand{\thefootnote}{\fnsymbol{footnote}}
\footnotetext[1]{Equal contribution.}
\footnotetext[2]{Corresponding authors: 
Bin Cao (\href{mailto:bincao@hkust-gz.edu.cn}{bincao@hkust-gz.edu.cn}) and  
Qiang Liu (\href{mailto:qiang.liu@nlpr.ia.ac.cn}{qiang.liu@nlpr.ia.ac.cn}).}
\endgroup

\begin{abstract}
Multiphase powder X-ray diffraction (PXRD) analysis remains a fundamental bottleneck in structure identification, as real-world synthesis often produces complex mixtures whose constituent phases (components) cannot be reliably disentangled. While recent advances in representation-based crystal retrieval and generation suggest the possibility of inferring structures directly from PXRD, existing approaches largely assume single-phase inputs and break down in multiphase settings.
Here, we present XDecomposer, a prior-free framework for joint decomposition and identification of multiphase XRD patterns without requiring candidate phase lists, structural templates, or prior knowledge of phase number. We formulate multiphase diffraction analysis as a set prediction problem, where the model infers an unordered set of phase-resolved components, their mixture proportions, and corresponding structural representations within a unified architecture. A phase-query-driven decomposition mechanism, together with diffraction-consistent physical reconstruction, enables accurate source separation while preserving crystallographic fidelity.
Extensive experiments on both simulated and experimental datasets show that XDecomposer substantially improves reconstruction accuracy and phase identification across diverse chemical systems, while maintaining strong generalization to unseen mixtures. These results provide a practical route toward data-driven, source-resolved multiphase XRD analysis and reduce long-standing dependence on prior-guided iteratively phase matching. The code is openly available at 
\url{https://github.com/Licht0812/XDecomposer}
\end{abstract}

\section{Introduction}

Powder X-ray diffraction (PXRD) is a foundational technique for determining crystal structure \cite{dinnebier2015powder,bunaciu2015x,young1993rietveldmethod,cao2026physics}. In a typical PXRD workflow, phase identification provides the basis for downstream structural refinement and quantitative analysis. Widely used refinement methods, such as Rietveld \cite{rietveld1969}, Pawley \cite{pawley1981}, and Le Bail \cite{le1988ab}, generally assume that the constituent phases are known a priori; when this assumption is violated, subsequent analyses can become unreliable. Phase identification is relatively well established for single-phase samples \cite{lutterotti2019full,xqueryer2025}. However, in practice, samples often contain multiple crystalline phases whose diffraction signals are superimposed in a single one-dimensional pattern. In such multiphase settings, the goal is no longer to identify a single phase, but to simultaneously determine all constituent phases and estimate their relative abundances. Unknown compositions, severe peak overlap, and experimental distortions further complicate this task \cite{bin2025simxrd,maffettone2021xca,szymanski2023adaptively,schuetzke2024fastscreen}. As a result, multiphase phase identification has become a central bottleneck in PXRD analysis.

From a machine learning perspective, single-phase identification can be cast as mapping a one-dimensional PXRD pattern to a single phase or structural representation. This setting has been widely studied using classification \cite{xqueryer2025,bin2025simxrd,maffettone2021xca,salgado2023automated,lee2020deep,le2023deep,szymanski2021probabilistic}, retrieval \cite{chen2024crystal,lai2025end,xu2026kan,zhang2025structural}, and generative modeling \cite{li2025powder,yu2026equivariant,feng2025interpreting,guo2025ab}. By contrast, multiphase analysis is not simply a multi-label extension of single-phase identification. A mixed pattern is generated by multiple latent diffraction components whose identities, number, and proportions are generally unknown. This makes the task more naturally formulated as a single-channel blind source separation (BSS) problem: given a mixed PXRD pattern, a model must jointly identify the constituent phases, reconstruct the diffraction signal of each phase, and estimate their relative proportions. This formulation highlights the underdetermined nature of the problem, in which multiple unobserved components must be recovered from a single observed signal, often without component-level supervision (see Appendix~\ref{xrd_sim}).

A natural approach is to exploit the additive structure of diffraction signals. Existing methods therefore adopt a sequential identification paradigm \cite{szymanski2023adaptively,mdi-jade}: they iteratively identify a dominant phase, subtract its contribution, and repeat on the residual. However, this strategy suffers from a fundamental limitation: phase proportions are unknown, making each subtraction inherently inaccurate. Moreover, peak overlap causes subtraction to distort signals from other phases, leading to error accumulation across iterations. As a result, even strong single-phase models often fail in multiphase settings, revealing the inadequacy of sequential identification and motivating the need for a new paradigm.

Here, we formulate prior-free multiphase XRD analysis as a blind source separation problem, in which a model directly decomposes a mixed diffraction pattern into an unordered set of single-phase components and their proportions. Existing approaches either rely on predefined phase sets or sequential subtraction, which limits their ability to jointly recover all components.
To this end, we propose XDecomposer for direct whole-pattern decomposition. It is trained in two stages: self-supervised pretraining on large-scale single-phase data to learn general diffraction patterns, followed by supervised training on multiphase mixtures for component-level decomposition. XDecomposer integrates convolutional encoding for local peak features, Transformer-based global context modeling, a query-based mechanism to separate individual phases, and physically constrained reconstruction.
Experiments on simulated and real-word experimental datasets show that XDecomposer accurately decomposes mixtures, reconstructs phase-specific signals, and estimates phase proportions, establishing a new perspective for multiphase PXRD analysis. The contributions are as follows:

\begin{itemize}[leftmargin=*]
    \item \textbf{Problem formulation.} We cast multiphase PXRD analysis as a permutation-invariant set prediction problem under unknown phase cardinality, providing an alternative to sequential identification and enabling direct, prior-free decomposition of mixture patterns.

    \item \textbf{Modeling approach.} We introduce XDecomposer, a query-based decomposition model that combines global context modeling with phase-conditioned latent modulation to separate phase-specific components from a single mixed diffraction pattern.

    \item \textbf{Physics-aware reconstruction.} We incorporate physically motivated constraints, including non-negativity and mixture consistency, through a mask-based reconstruction mechanism, improving stability and plausibility of the recovered components.

    \item \textbf{Empirical evaluation.} We conduct extensive experiments on both simulated and real-world datasets, showing that XDecomposer consistently outperforms strong baselines across multiple metrics and exhibits robust generalization to unseen mixtures.
\end{itemize}

\section{Related Work}
\label{sec:Related_work}
\paragraph{Query strategy}

For single-phase materials, identification typically matches measured diffraction patterns to reference entries, such as the Powder Diffraction File, using peak positions and relative intensities \cite{mdi-jade,cao2026physics} (see Appendix~\ref{SM_method}). 
For multiphase systems, the observed pattern is treated as a superposition of phase signals and analyzed sequentially by identifying a dominant phase, subtracting its contribution, and repeating on the residual. 
Such methods therefore cannot jointly infer all mixture components and, despite their effectiveness in controlled settings, depend strongly on candidate-phase priors and are sensitive to peak overlap, noise, and incomplete or biased reference databases.

\paragraph{Learning Strategies}

Recent advances leverage machine learning to automate phase identification from PXRD patterns. For single-phase materials, the problem is well-studied and typically formulated as classification, retrieval, or generation. Classification methods formulate phase identification as a supervised multi-class classification problem, directly assigning an input diffraction pattern to one of the predefined phase labels in the training database \cite{xqueryer2025,bin2025simxrd,maffettone2021xca,salgado2023automated,lee2020deep,le2023deep,szymanski2021probabilistic}. Retrieval-based approaches, in contrast, learn feature representations that enable similarity search, ranking candidate phases by matching a query pattern against a database of known phases rather than directly predicting a class label \cite{chen2024crystal,lai2025end,xu2026kan,zhang2025structural}. Generation methods model the inverse diffraction process, reconstructing crystal structures from PXRD patterns to facilitate phase identification \cite{li2025powder,yu2026equivariant,feng2025interpreting,guo2025ab}.

In contrast, multiphase analysis remains underexplored. Existing methods typically extend these formulations to multi-label classification \cite{lee2020deep,le2023deep,fei2026dara,greasley2023exploring} or regression \cite{alade2022prediction,alfares2025machine}, often with fixed candidate phase sets, or adopt sequential strategies that iteratively identify and subtract phase contributions from the mixture \cite{szymanski2023adaptively}.

However, these methods remain fundamentally limited: predefined phase sets and labeled data restrict generalization to unseen mixtures; sequential subtraction accumulates errors under severe peak overlap; and generation-based methods may produce physically inconsistent structures. Consequently, current approaches struggle to reliably disentangle and recover individual phase components from multiphase PXRD patterns.

\paragraph{Blind Source Separation}
We relate multiphase PXRD decomposition to BSS, set prediction, and self-supervised representation learning. In BSS, permutation invariant training (PIT) addresses output permutation ambiguity \cite{yu2017pit,kolbaek2017upit}, while Conv-TasNet \cite{luo2019convtasnet} and SepFormer \cite{subakan2021sepformer} demonstrate the effectiveness of mask-based reconstruction and Transformer-based long-range modeling. In set prediction, Deep Sets \cite{zaheer2017deepsets}, Set Transformer \cite{lee2019settransformer}, and DETR \cite{carion2020detr} establish permutation-invariant and query-based prediction paradigms. In representation learning, Transformer \cite{vaswani2017transformer}, Masked Autoencoder \cite{he2022mae}, and FiLM \cite{perez2018film} support global modeling, self-supervised pretraining, and conditional feature modulation. However, multiphase PXRD decomposition is not a generic source separation problem: diffraction patterns are governed by crystal-structure-dependent peak positions and intensities constrained by Bragg’s law, while severe peak overlap makes phase-wise disentanglement highly ill-posed. Moreover, the target is an unordered set of constituent phases with variable cardinality rather than a fixed set of sources. Existing BSS and set prediction methods do not explicitly model these physics-driven constraints, limiting their effectiveness for physically consistent PXRD decomposition.

In summary, existing approaches, whether database-driven, learning-based, or adapted from generic BSS frameworks, fail to provide a prior-free, physically consistent, and unified solution for jointly decomposing and identifying multiphase PXRD patterns.

\section{Preliminaries}
\label{sec:Prelim}
\subsection{Phase Identification}

In X-ray diffraction analysis, diffraction patterns encode crystal structures through peak positions and intensities. For a single-phase material, the observed pattern can be approximated as a deterministic function of its crystal structure, including lattice parameters, atomic positions, and chemical elements. In practice, extrinsic factors such as sample quality, preferred orientation, and instrument conditions introduce distortions and reduce phase distinguishability (see Appendix~\ref{xrd_sim}). A single-phase pattern can generally be represented as a function $x(\theta)$ over the diffraction angle $\theta$.

For multiphase materials, the observed pattern is a superposition of single-phase signals governed by diffraction physics \cite{pecharsky2003fundamentals}. Given $K$ constituent phases, the measured pattern $x(\theta)$ can be expressed as
\begin{equation}
x(\theta) = \sum_{k=1}^{K} w_k \, y_k(\theta) + \epsilon(\theta),
\label{eq:mixture}
\end{equation}
where $y_k(\theta)$ denotes the diffraction pattern of the $k$-th phase, $w_k$ is its mixture coefficient, and $\epsilon(\theta)$ accounts for noise and trace impurities, whose contribution is negligible in multiphase decomposition. Recovering both $\{y_k\}_{k=1}^{K}$ and $\{w_k\}_{k=1}^{K}$ from a single observation $x(\theta)$ is the inherent mission of multiphase identification. 
Existing approaches typically restrict the solution space to predefined candidate phases \cite{lee2020deep,le2023deep,greasley2023exploring} and rely on iterative fitting \cite{szymanski2023adaptively,ito2026enabling}. In contrast, we address the realistic prior-free setting, where both the number of phases and their diffraction components are unknown and must be inferred directly from the observed mixture.

Novel crystal structure determination from multiphase diffraction data requires two sequential stages: mixture decomposition and crystal structure identification. The latter comprises both in-library and out-of-library settings (Appendix~\ref{app:In-out-tasks}). Combined with database matching, XDecomposer completes the full pipeline for in-library identification, whereas out-of-library identification requires integration with generative crystal structure prediction methods and remains an open research problem.

\subsection{Problem Definition}
\label{subsec:problem}

Let the input pattern be sampled on a discrete $2\theta$ grid of length $L$. After standard preprocessing, the mixed diffraction pattern is decomposed as
\begin{equation}
\mathcal{D}\bigl(x(\theta)\bigr)
=
\left\{ \hat{y}_k \right\}_{k=1}^{K},
\qquad
\text{subject to}
\qquad
x(\theta) \approx \sum_{k=1}^{K} \hat{y}_k,
\quad
1 \leq K \leq K_{\max},
\label{eq:decomposition}
\end{equation}

where $\mathcal{D}(\cdot)$ maps the mixture to a set of decomposed single-phase contributions. Each $\hat{y}_k  \approx w_k \, y_k(\theta)\in \mathbb{R}_+^{L}$ denotes the estimated contribution of the $k$-th phase, incorporating realistic experimental effects and phase proportions. Each component is assigned a confidence score $p_k$, and active phases are selected by thresholding with $\tau$, yielding $\hat{\mathcal{Y}}^{*} = \{ \hat{y}_k \mid p_k > \tau \}$.

\subsection{Data Synthesis}
\label{sec:data_synthesis}
Under standard diffraction conditions, a mixed PXRD pattern can be approximated as a linear superposition of its constituent phase signals, as shown in Eq.~\ref{eq:mixture}. We first construct simulated and experimental single-phase PXRD libraries, and then generate mixed PXRD data using a multiphase mixing strategy. Details of single-phase data generation and processing are given in Appendices~\ref{app:sim_gen} and~\ref{app:rruff}, and the mixing strategy is described in Appendix~\ref{app:mixing}.

\textbf{Simulated data}: We parse 100,315 crystal structures from the Materials Project database~\cite{jain2013materialsproject} (MP500) and generate 20 single-phase PXRD patterns per structure using PySimXRD~\cite{bin2025simxrd}, yielding $100{,}315 \times 20$ simulated patterns. These patterns are then mixed with phase fractions $\{w_k\}$ and normalized to $\max_{\theta} x(\theta)=1$. Simulations are performed on a high-performance computing cluster with perturbations in peak position, peak width, background, and instrumental geometry.

\textbf{Experimental data}: Real single-phase patterns are first corrected by WPEM~\cite{cao2026ai} to remove baseline and background, and are then mixed using controllable phase fractions. The mixtures are generated on the fly during data loading to increase sample diversity.

\section{Methodology}
\label{sec:method}

\subsection{Overall Framework}

\begin{figure}[htbp]
	\centering
	\includegraphics[width=\linewidth]{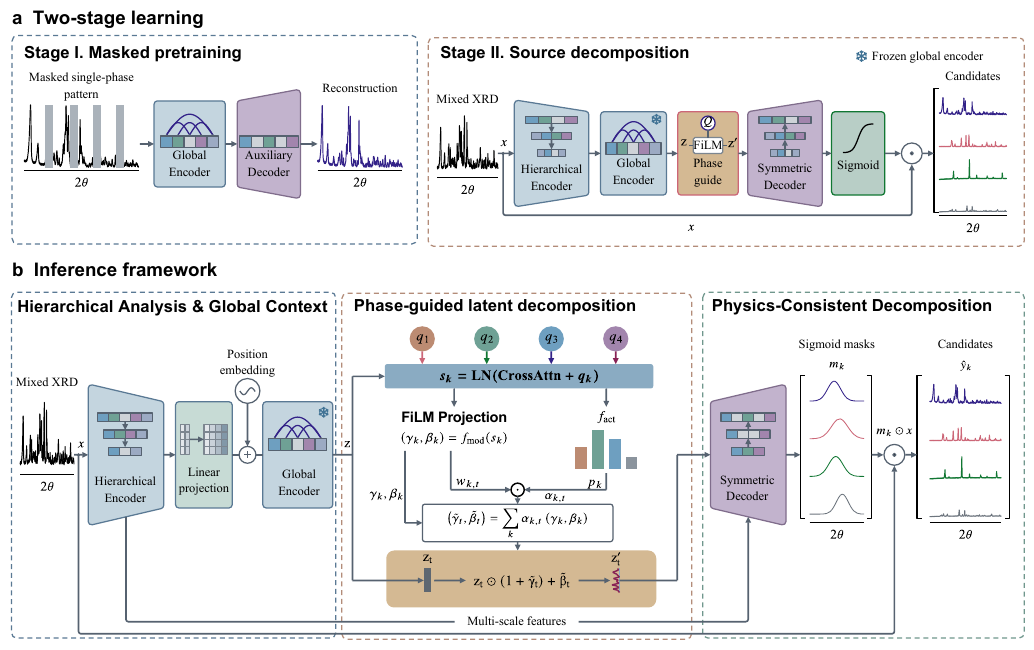}
	\caption{Overall framework of XDecomposer.
    \textbf{Part I} shows the two-stage training strategy: masked-reconstruction pretraining of the global-context encoder $G(\cdot)$ on single-phase patterns, followed by mixture training with $G(\cdot)$ frozen.  \textbf{Part II} presents the inference framework including \textbf{a}, hierarchical analysis and global context modeling; \textbf{b}, phase-guided latent decomposition; and \textbf{c}, physics-consistent reconstruction.}
	\label{fig:arch}
\end{figure}

We formulate prior-free multiphase XRD decomposition as a set-prediction-based one-dimensional BSS problem. XDecomposer uses a pretrained global bottleneck to capture transferable peak structure and is trained end-to-end with permutation-invariant separation loss, slot-activity supervision, and mixture-consistency regularization (Fig.~\ref{fig:arch}). It enables one-shot inference directly from mixed patterns without candidate phase lists, structural templates, or database retrieval.
Given a preprocessed mixed diffraction pattern $x \in \mathbb{R}_{+}^{L}$, the model predicts a fixed-size set of $K_{\max}=4$ output slots,
\[
\hat{\mathcal{Y}} = \{(\hat{y}_k, p_k)\}_{k=1}^{K_{\max}},
\]
where $\hat{y}_k \in \mathbb{R}_{+}^{L}$ denotes the $k$-th candidate contribution and $p_k \in [0,1]$ represents its activity probability. During training, the input is a multiphase mixture $x$, and the supervision target is the set of constituent single-phase components, zero-padded to four slots when fewer than four phases are present. During inference, inactive slots are removed according to their activity scores, yielding a variable-size set of active components.

\subsection{Module}
\label{subsec:arch}

\paragraph{Hierarchical analysis and global context modeling.}
As shown in Fig.~\ref{fig:arch}\textbf{a}, the input mixed diffraction pattern is first processed by a hierarchical one-dimensional convolutional analyzer $E(\cdot)$ to extract multi-scale local spectral features, yielding a latent representation $h=E(x)\in\mathbb{R}^{T\times D}$, where $T$ is the latent sequence length and $D$ is the feature dimension. The shallow convolutional layers capture local peak shapes, peak-width variations, and intensity perturbations. The deeper layers enlarge the receptive field and encode mid-range contextual relations among neighboring peak groups. Intermediate-scale features are further retained and fused during reconstruction to preserve the resolution of peak positions and line shapes.

The latent features $h$ are then projected to the target dimension, combined with positional encodings, and passed to a global context encoder $G(\cdot)$, which produces a shared latent representation
$z=G(h)\in\mathbb{R}^{T\times D}$
This global context modeling is motivated by the observation that multiple Bragg reflections from the same phase exhibit long-range dependencies in both peak positions and relative intensities. These dependencies are jointly governed by structure factors and symmetry~\cite{kaduk2021powder,stephens2019physics}. Consequently, purely local convolutions are insufficient to recover phase-consistent structural patterns.

\paragraph{Phase-guided latent decomposition.}
Fig.~\ref{fig:arch}\textbf{b} illustrates the phase-guided latent decomposition module, which decomposes the shared latent space into phase-conditioned responses. We introduce a set of learnable phase queries $Q=\{q_k\}_{k=1}^{K_{\max}}$. For each slot $k$, $q_k$ attends to the shared latent sequence $z$ through cross-attention:
\begin{equation}
	s_k=\mathrm{CrossAttn}(q_k,z,z), \qquad k=1,\dots,K_{\max}.
	\label{eq:cross_attn}
\end{equation}
The slot summary $s_k$ is projected to an activity logit $a_k=f_{\mathrm{act}}(s_k)$ and FiLM modulation parameters $(\gamma_k,\beta_k)=f_{\mathrm{mod}}(s_k)$, with activity probability $p_k=\sigma(a_k)$.

To obtain stable and complementary slot assignments, we combine position-wise spatial competition with sample-level activity gating. At latent position $t$, the spatial competition weight is computed from the similarity between the slot modulation vector $\gamma_k$ and the local representation $z_t$:
\begin{equation}
	w_{k,t}
	=
	\frac{
		\exp\!\big(\langle \gamma_k,z_t\rangle/\sqrt{D}\big)
	}{
		\sum_{j=1}^{K_{\max}}
		\exp\!\big(\langle \gamma_j,z_t\rangle/\sqrt{D}\big)
	},
	\label{eq:spatial_compete}
\end{equation}
where $\sum_k w_{k,t}=1$. The effective slot weight is then defined as
$\alpha_{k,t}=w_{k,t}p_k$, where $w_{k,t}$ enforces local competition and $p_k$ controls whether slot $k$ is active for the current sample.

The effective weights are used to aggregate slot-specific FiLM parameters \cite{perez2018film} into spatially adaptive modulation terms:
\begin{equation}
	\tilde{\gamma}_t=\sum_k \alpha_{k,t}\gamma_k,
	\qquad
	\tilde{\beta}_t=\sum_k \alpha_{k,t}\beta_k.
\end{equation}
The shared latent representation is then modulated by feature-wise affine transformation:
\begin{equation}
	z'_t=z_t\odot\big(1+\tilde{\gamma}_t\big)+\tilde{\beta}_t
	\label{eq:film_modulate}
\end{equation}
Equivalently, the modulation can be viewed as a weighted combination of slot-conditioned responses around the shared latent feature, where each slot response is defined as
$z_t^{(k)}=z_t\odot(1+\gamma_k)+\beta_k$. This formulation performs phase decomposition directly in the shared latent space, before decoding, while preserving the feature-wise affine structure of FiLM \cite{perez2018film}. The aggregated representation $z'$ is then passed to the reconstruction module together with the retained multi-scale features.

\paragraph{Physics-consistent reconstruction.}
Fig.~\ref{fig:arch}\textbf{c} shows the physics-consistent reconstruction module. The decoder takes the modulated latent representation $z'$ together with the retained multi-scale features and predicts $K_{\max}$ soft masks over the full angular domain. Denoting the decoder output by $U=D(z',\mathcal{S})\in\mathbb{R}^{K_{\max}\times L}$, where $\mathcal{S}$ represents the retained multi-scale features, the soft mask associated with slot $k$ is defined as $m_k=\sigma(U_k)\in(0,1)^L$. The model then obtains the corresponding single-phase contribution by point-wise multiplication:
\begin{equation}
	\hat{y}_k=m_k\odot x.
	\label{eq:mask_recon}
\end{equation}
This mask-based parameterization is similar in form to mask-based separation methods in single-channel source separation. In the present setting, however, it is further constrained by the physical properties of the diffraction pattern, especially non-negativity and mixture consistency \cite{luo2019convtasnet,subakan2021sepformer}. Since the input diffraction pattern is non-negative and $m_k\in(0,1)$, the model naturally satisfies $0\le \hat{y}_k\le x$. As a result, each output remains non-negative and cannot exceed the input intensity at any position. This property helps suppress spurious peak responses, especially in low-background regions. The reconstructed mixed pattern is obtained by summing all candidate single-phase contributions, i.e.,
$\hat{x}=\sum_{k=1}^{K_{\max}}\hat{y}_k$.
In this way, part of the mixture-consistency constraint is built directly into the output parameterization, rather than being enforced only through a posterior loss term.

\subsection{Training and Inference}
\label{subsec:train_infer}

XDecomposer adopts a two-stage training strategy. \textbf{In Stage I}, an MAE-style masked reconstruction task \cite{he2022mae} is constructed on single-phase diffraction patterns to pretrain the global-context encoder $G(\cdot)$ for transferable pure-phase structural representations. An auxiliary decoder, used only during pretraining, provides reconstruction supervision.\textbf{In Stage II}, the pretrained global-context encoder is transferred to the full decomposition model and kept frozen, while only the decomposition-related modules are optimized. This allows the model to learn single-phase separation, slot activity prediction, and mixture consistency from multiphase mixtures.
Since the output slots are unordered, decomposition training uses permutation-invariant training (PIT) \cite{yu2017pit,kolbaek2017upit}. Let $\{y_k\}_{k=1}^{K}$ denote the ground-truth single-phase contributions and $\{\hat{y}_k\}_{k=1}^{K_{\max}}$ the predicted set. The optimal assignment is
\begin{equation}
	\pi^{*}
	=
	\arg\min_{\pi}
	\sum_{k=1}^{K}\mathcal{L}_{\mathrm{sep}}(\hat{y}_{\pi(k)},y_k).
	\label{eq:pit}
\end{equation}
Based on this assignment, the overall training objective is
\begin{equation}
	\mathcal{L}
	=
	\mathcal{L}_{\mathrm{sep}}^{\mathrm{PIT}}
	+
	\lambda_{\mathrm{act}}\mathcal{L}_{\mathrm{act}}
	+
	\lambda_{\mathrm{mix}}\mathcal{L}_{\mathrm{mix}},
	\label{eq:total_loss}
\end{equation}
where $\mathcal{L}_{\mathrm{sep}}^{\mathrm{PIT}}$ is the PIT-aligned separation loss, $\mathcal{L}_{\mathrm{act}}$ supervises slot activity, and $\mathcal{L}_{\mathrm{mix}}$ enforces consistency between the reconstructed mixture and the input pattern. The separation loss includes amplitude, shape, and geometric terms, with full definitions provided in Appendix~\ref{app:loss}. \textbf{At inference}, XDecomposer decomposes the input pattern in a single forward pass and outputs separated components with activity scores.

\section{Experiments}
\label{sec:exp}

\subsection{Benchmarks}

\paragraph{Datasets} Two large-scale benchmarks are employed in this work. \textbf{Simulated data} use the corpus described in Section~\ref{sec:data_synthesis}. The dataset is split into training, validation, and test sets with an 8:1:1 ratio, ensuring no structural overlap across subsets (see Appendices~\ref{app:mixing} and~\ref{app:split}).
\textbf{Experimental data} consist of 662 measured single-phase PXRD patterns from the RRUFF database~\cite{rruff2015}. After baseline correction and background removal, mixed patterns are generated with controllable phase fractions. Model performance is evaluated using five-fold cross-validation (see Appendix~\ref{app:rruff}).

\paragraph{Baselines}
While related methods exist in BSS and PXRD analysis, none directly address prior-free, physically consistent decomposition under unknown phase cardinality (see Section~\ref{sec:Related_work}). To ensure comprehensive evaluation, we include both domain-specific PXRD models and strong general-purpose sequence models that have demonstrated competitive performance on diffraction data.

\begin{itemize}[leftmargin=*,itemsep=1pt,topsep=1pt,parsep=0pt,partopsep=0pt]
    \item \textbf{Domain Models}: We include three representative domain-specific models developed for PXRD phase identification. XQueryer \cite{xqueryer2025} is the first cross-chemical general model for single-phase identification based on classification. XRD\_Proportion\_Inference \cite{simonnet2024phasequant,simonnet2025vitxrd} employs Vision Transformer (ViT) and CNN architectures to predict phase proportions, effectively performing regression over mixture ratios for a fixed set of candidate phases. XRDAutoAnalyzer \cite{szymanski2021probabilistic,szymanski2024xrdpdf} is a classification-based model for single-phase identification; its multi-phase extension adopts a sequential phase identification strategy.
    
    \item \textbf{Sequence Models}: We further include strong general-purpose sequence models, including Transformer \cite{vaswani2017transformer}, iTransformer \cite{DBLP:journals/corr/abs-2310-06625}, and PatchTST \cite{DBLP:conf/iclr/NieNSK23}. These models have shown competitive performance on PXRD-related tasks \cite{xqueryer2025, bin2025simxrd} and represent a class of flexible, domain-agnostic approaches capable of modeling complex diffraction patterns.
\end{itemize}

Baseline configurations differ between simulated and experimental settings. For simulated data, we report all baselines. For experimental data, XRDAutoAnalyzer is excluded because it is limited to fixed mixture component sets and its performance collapses on the RRUFF experimental dataset. We evaluate all models under a unified protocol to assess the extent to which their outputs can recover single-phase representations from mixture observations. Implementation details of all baselines are provided in Appendix~\ref{app:baselines}. Model size, inference cost, and deployment overhead are summarized in Table~\ref{tab:training_efficiency} of Appendix~\ref{app:impl}.

\paragraph{Evaluation Metrics}
We report five evaluation metrics: the Pearson correlation coefficient (unitless), mean peak-position deviation $\overline{\Delta 2\theta}$ (in $^\circ$), full width at half maximum error $\Delta\mathrm{FWHM}$ (in $^\circ$), Top-1 retrieval accuracy (\%), and Top-10 retrieval accuracy (\%). These metrics evaluate global pattern-shape similarity, local peak-position accuracy, line-width fidelity, and downstream phase retrieval discriminability, respectively. Formal definitions are provided in Appendix~\ref{app:metrics}.

\begin{table}[h]
	\centering
	\caption{Baseline comparison on simulated PXRD mixtures with $K=2,3,4$.}
	\label{tab:main_results}
    \renewcommand{\arraystretch}{1}
    \setlength{\tabcolsep}{4pt}
	\resizebox{0.9\columnwidth}{!}{%
		\begin{tabular}{c l p{3.7cm} c c c c c}
			\toprule
			$K$ & Method Type & Model & Pearson $\uparrow$ & Peak Shift ($^\circ$) $\downarrow$ & $\Delta$FWHM ($^\circ$) $\downarrow$ & Top-1 (\%) $\uparrow$ & Top-10 (\%) $\uparrow$ \\
			\midrule
			
			\multirow{7}{*}{2}
			& \multirow{3}{*}{Domain}
			& XQueryer & 0.6873 & 0.0363 & 5.0005 & N/A & N/A \\
			& & XRD\_Proportion\_Inference & 0.6330 & 0.0742 & 10.7957 & 30.44 & 43.90 \\
			& & XRDAutoAnalyzer & 0.0135 & 0.0897 & 27.5939 & 1.62 & 2.43 \\
			\cmidrule(lr){2-8}
			& \multirow{3}{*}{Sequence}
			& Transformer & \underline{0.9486} & \underline{0.0322} & \underline{3.1322} & \underline{76.43} & \underline{88.17} \\
			& & iTransformer & 0.9291 & 0.0347 & 3.6935 & 74.25 & 86.42 \\
			& & PatchTST & 0.9278 & 0.0336 & 3.3702 & 75.29 & 86.74 \\
			\cmidrule(lr){2-8}
			& Proposed & \textbf{XDecomposer} & \textbf{0.9691} & \textbf{0.0272} & \textbf{2.4330} & \textbf{87.92} & \textbf{92.50} \\
			
			\midrule
			\multirow{7}{*}{3}
			& \multirow{3}{*}{Domain}
			& XQueryer & 0.5943 & 0.0501 & 8.6083 & N/A & N/A \\
			& & XRD\_Proportion\_Inference & 0.5274 & 0.0773 & 12.2785 & 13.95 & 23.35 \\
			& & XRDAutoAnalyzer & 0.0043 & 0.0900 & 26.6406 & 0.52 & 0.81 \\
			\cmidrule(lr){2-8}
			& \multirow{3}{*}{Sequence}
			& Transformer & \underline{0.8608} & \underline{0.0414} & 4.4464 & \underline{64.60} & \underline{78.62} \\
			& & iTransformer & 0.7949 & 0.0444 & 4.9595 & 58.19 & 72.45 \\
			& & PatchTST & 0.8010 & 0.0429 & \underline{4.4322} & 60.21 & 73.94 \\
			\cmidrule(lr){2-8}
			& Proposed & \textbf{XDecomposer} & \textbf{0.9101} & \textbf{0.0389} & \textbf{3.7836} & \textbf{81.13} & \textbf{86.81} \\
            
			\midrule
			\multirow{7}{*}{4}
			& \multirow{3}{*}{Domain}
			& XQueryer & 0.5372 & 0.0612 & 12.9371 & N/A & N/A \\
			& & XRD\_Proportion\_Inference & 0.4092 & 0.0799 & 13.2168 & 5.68 & 11.25 \\
			& & XRDAutoAnalyzer & 0.0011 & 0.0909 & 25.5976 & 0.14 & 0.27 \\
			\cmidrule(lr){2-8}
			& \multirow{3}{*}{Sequence}
			& Transformer & \underline{0.7491} & \underline{0.0499} & 5.7736 & \underline{48.41} & \underline{62.13} \\
			& & iTransformer & 0.7084 & 0.0527 & 6.4041 & 40.54 & 53.52 \\
			& & PatchTST & 0.7146 & 0.0519 & \underline{5.7424} & 42.66 & 55.81 \\
			\cmidrule(lr){2-8}
			& Proposed & \textbf{XDecomposer} & \textbf{0.8322} & \textbf{0.0480} & \textbf{5.2410} & \textbf{69.68} & \textbf{76.21} \\
			\bottomrule
		\end{tabular}%
	}
    \begin{minipage}{\textwidth}
		\footnotesize
		\textit{Note:} Top-k measures whether the ground-truth phase appears within the top k retrieved candidates at the phase level. XQueryer is not included in the Top-$k$ comparison because its framework does not support outputting retrieval results required for computing Top-$k$ scores. The profile factors, $R_p$ and $R_{wp}$, are reported in Table~\ref{tab:profile_factors}.
	\end{minipage}
\end{table}

\subsection{Results}
\begin{figure*}[h]
	\centering
	\includegraphics[width=0.95\linewidth]{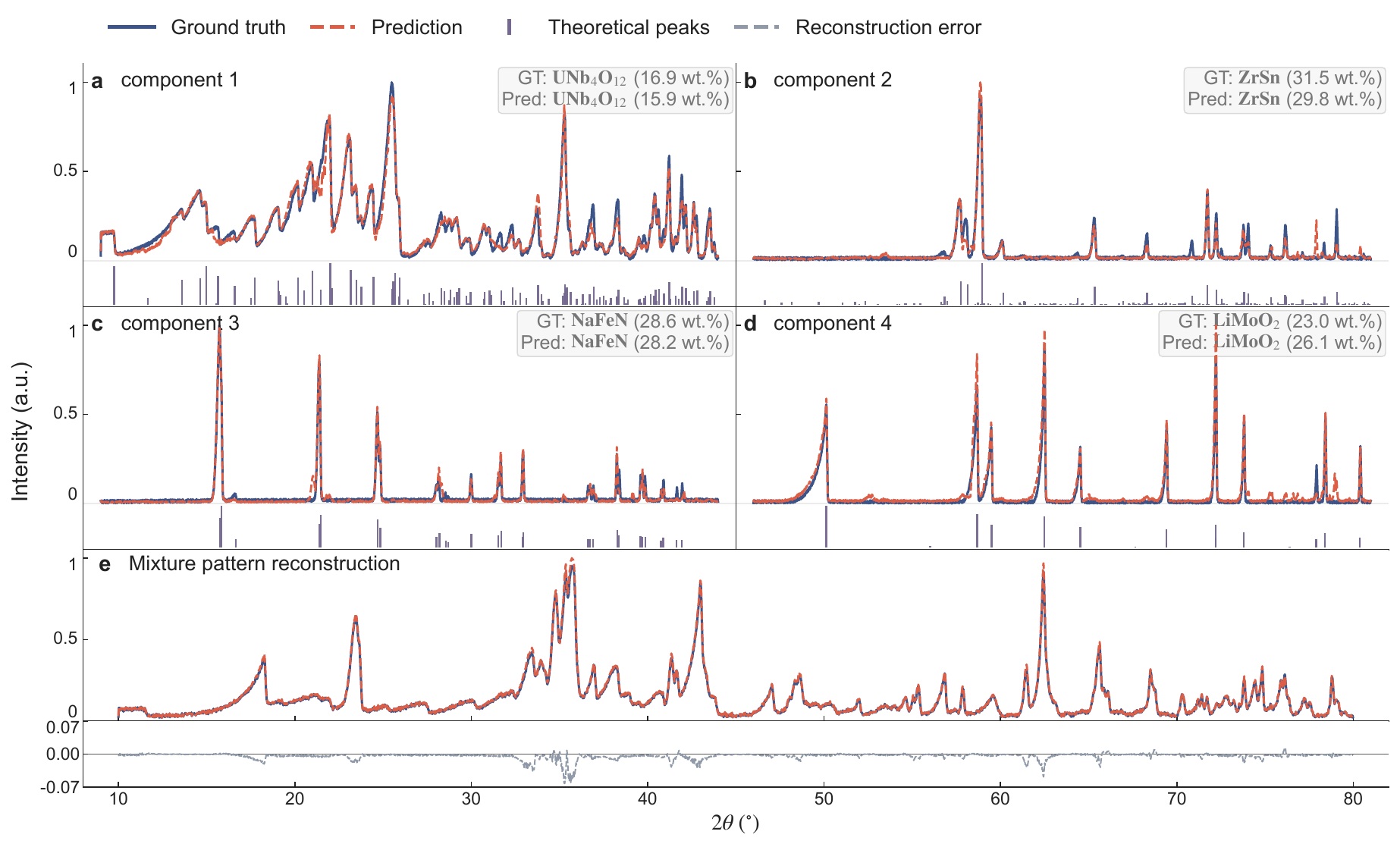}
	\caption{Representative quantitative results. \textbf{a}--\textbf{d}: true single-phase contributions, predicted patterns, and theoretical peak positions. \textbf{e}: mixture reconstruction consistency.}
	\label{fig:case_studies}
\end{figure*}

\paragraph{Case Study} Figure~\ref{fig:case_studies} presents a representative four-phase whole pattern decomposition example. XDecomposer accurately reconstructs the overall diffraction pattern while correctly identifying the contribution of each source phase, even in the most challenging cases with severe peak broadening. Additional quantitative cases for binary, ternary, and quaternary mixtures are provided in Appendix~\ref{app:qualitative}, Figs.~\ref{fig:sim_k2}, \ref{fig:sim_k3}, and \ref{fig:sim_k4}. These results empirically indicate that XDecomposer achieves accurate whole-pattern decomposition with \textbf{near-refinement-level quality} \cite{cao2026ai}.

\paragraph{Simulated Data}
Table~\ref{tab:main_results} reports results on simulated data for $K=2,3,4$. XDecomposer achieves the best overall performance across all settings. Among methods supporting Top-$k$ retrieval evaluation, it consistently outperforms baselines in Pearson correlation, Peak Shift, $\Delta$FWHM, Top-1, and Top-10, indicating more accurate recovery of overall shape, peak positions, and line profiles, and consequently better downstream phase retrieval.

As $K$ increases from 2 to 4, all methods degrade due to stronger peak overlap and greater phase ambiguity. In contrast, XDecomposer degrades more gracefully, showing greater robustness in complex multiphase scenarios due to large-scale XRD pretraining. Although prior domain-specific models remain competitive with general sequence models on simulated data, their strong single-phase priors limit multiphase decomposition, especially in spectral fidelity, peak-geometry consistency, and retrieval accuracy.

\begin{table}[htbp]
	\centering
	\caption{Baseline comparison on RRUFF real data with $K=2,3,4$ (five-fold cross-validation).}
	\label{tab:exp_results}
	\renewcommand{\arraystretch}{1}
	\setlength{\tabcolsep}{3pt}
	\resizebox{0.95\columnwidth}{!}{%
		\begin{tabular}{c c l c c c c c}
			\toprule
			$K$ & Method Type & Model & Pearson $\uparrow$ & Peak Shift ($^\circ$) $\downarrow$ & $\Delta$FWHM ($^\circ$) $\downarrow$ & Top-1 (\%) $\uparrow$ & Top-10 (\%) $\uparrow$ \\
			\midrule
			\multirow{6}{*}{2}
			& \multirow{2}{*}{Domain}
			& XQueryer & \underline{0.6519 $\pm$ 0.0244} & \textbf{0.0164 $\pm$ 0.0006} & \textbf{1.0097 $\pm$ 0.0842} & \multicolumn{1}{c}{N/A} & \multicolumn{1}{c}{N/A} \\
			& & XRD\_Proportion\_Inference & 0.4808 $\pm$ 0.1448 & 0.0652 $\pm$ 0.0160 & 17.2678 $\pm$ 9.2924 & 44.60 $\pm$ 18.44 & 75.29 $\pm$ 19.06 \\
			\cmidrule(lr){2-8}
			& \multirow{3}{*}{Sequence}
			& Transformer & 0.5117 $\pm$ 0.0114 & 0.0245 $\pm$ 0.0011 & 2.7504 $\pm$ 0.2616 & 40.68 $\pm$ 1.66 & 66.80 $\pm$ 2.32 \\
			& & iTransformer & 0.5984 $\pm$ 0.0150 & 0.0221 $\pm$ 0.0027 & 3.3806 $\pm$ 0.5232 & \underline{49.34 $\pm$ 3.28} & \underline{89.30 $\pm$ 4.36} \\
			& & PatchTST & 0.5544 $\pm$ 0.0099 & \underline{0.0183 $\pm$ 0.0014} & 2.7112 $\pm$ 0.4100 & 49.08 $\pm$ 0.89 & 76.75 $\pm$ 1.62 \\
			\cmidrule(lr){2-8}
			& Proposed & XDecomposer & \textbf{0.6956 $\pm$ 0.0091} & 0.0197 $\pm$ 0.0018 & \underline{1.4538 $\pm$ 0.1293} & \textbf{65.31 $\pm$ 2.62} & \textbf{97.51 $\pm$ 0.67} \\
			\midrule
			\multirow{6}{*}{3}
			& \multirow{2}{*}{Domain}
			& XQueryer & \underline{0.5294 $\pm$ 0.0012} & \textbf{0.0221 $\pm$ 0.0019} & \textbf{1.4311 $\pm$ 0.2081} & \multicolumn{1}{c}{N/A} & \multicolumn{1}{c}{N/A} \\
			& & XRD\_Proportion\_Inference & 0.3769 $\pm$ 0.1326 & 0.0695 $\pm$ 0.0137 & 18.1328 $\pm$ 8.4472 & 31.29 $\pm$ 14.22 & 63.61 $\pm$ 21.25 \\
			\cmidrule(lr){2-8}
			& \multirow{3}{*}{Sequence}
			& Transformer & 0.3304 $\pm$ 0.0134 & 0.0302 $\pm$ 0.0021 & 3.3232 $\pm$ 0.3127 & 26.49 $\pm$ 3.15 & 48.84 $\pm$ 2.57 \\
			& & iTransformer & 0.3961 $\pm$ 0.0063 & 0.0279 $\pm$ 0.0018 & 4.0545 $\pm$ 0.5033 & 34.42 $\pm$ 1.41 & \underline{64.65 $\pm$ 0.76} \\
			& & PatchTST & 0.3856 $\pm$ 0.0061 & \underline{0.0240 $\pm$ 0.0015} & 3.1564 $\pm$ 0.4406 & \underline{35.02 $\pm$ 1.92} & 61.78 $\pm$ 1.07 \\
			\cmidrule(lr){2-8}
			& Proposed & XDecomposer & \textbf{0.5821 $\pm$ 0.0070} & 0.0286 $\pm$ 0.0015 & \underline{2.0912 $\pm$ 0.0545} & \textbf{53.49 $\pm$ 1.48} & \textbf{92.96 $\pm$ 1.38} \\
			
			\midrule
			
			\multirow{6}{*}{4}
			& \multirow{2}{*}{Domain}
			& XQueryer & \underline{0.4651 $\pm$ 0.0017} & \textbf{0.0277 $\pm$ 0.0017} & \textbf{1.8599 $\pm$ 0.2528} & \multicolumn{1}{c}{N/A} & \multicolumn{1}{c}{N/A} \\
			& & XRD\_Proportion\_Inference & 0.3158 $\pm$ 0.1249 & 0.0733 $\pm$ 0.0115 & 19.3598 $\pm$ 7.6530 & 23.47 $\pm$ 11.26 & \underline{55.58 $\pm$ 20.95} \\
			\cmidrule(lr){2-8}
			& \multirow{3}{*}{Sequence}
			& Transformer & 0.2461 $\pm$ 0.0097 & 0.0342 $\pm$ 0.0017 & 3.9567 $\pm$ 0.3040 & 21.02 $\pm$ 1.92 & 38.86 $\pm$ 2.28 \\
			& & iTransformer & 0.2834 $\pm$ 0.0107 & 0.0328 $\pm$ 0.0016 & 4.7631 $\pm$ 0.4952 & 25.86 $\pm$ 2.41 & 50.34 $\pm$ 1.02 \\
			& & PatchTST & 0.2806 $\pm$ 0.0090 & \underline{0.0298 $\pm$ 0.0015} & 3.8275 $\pm$ 0.4475 & \underline{26.25 $\pm$ 1.59} & 49.74 $\pm$ 1.72 \\
			\cmidrule(lr){2-8}
			& Proposed & XDecomposer & \textbf{0.5009 $\pm$ 0.0074} & 0.0359 $\pm$ 0.0018 & \underline{2.6735 $\pm$ 0.1577} & \textbf{41.90 $\pm$ 1.74} & \textbf{87.36 $\pm$ 1.80} \\
			
			\bottomrule
		\end{tabular}%
	}
	\begin{minipage}{\columnwidth}
		\footnotesize
		\textit{Note:} Results are reported as mean $\pm$ standard deviation over five folds. Top-1 and Top-10 are reported only for methods that provide retrieval outputs; otherwise the entries are marked as N/A.
	\end{minipage}
\end{table}

\paragraph{Experimental Data}
We further evaluate generalization on the real-world RRUFF dataset, reporting all metrics as mean $\pm$ standard deviation over five-fold cross-validation. As shown in Table~\ref{tab:exp_results}, all methods drop substantially on experimental data, indicating a clear sim-to-real gap, as multiphase modeling is more fragile since all constituent phases jointly contribute and single-phase variations can significantly influence the overall multiphase pattern, as discussed in Appendices~\ref{app:data_analysis} and \ref{xrd_sim}.

Baselines show different strengths on experimental data. XQueryer achieves the best Peak Shift and $\Delta$FWHM, indicating better preservation of local peak positions and widths, but this local advantage does not translate into better overall decomposition or retrieval performance. Pearson correlation and Top-$k$ results show that existing domain-specific and general sequence models tend to absorb noise, background variations, and local measurement effects into their predictions, thereby degrading decomposition quality. This confirms that multiphase decomposition differs from single-phase identification, as it requires jointly modeling multiple component contributions rather than merely aligning local peaks. In contrast, XDecomposer achieves superior or competitive results across metrics and maintains relatively small standard deviations under five-fold cross-validation, demonstrating its robustness and generalization on real experimental data.

\paragraph{Ablation Study.}
We conduct ablation studies on simulated data with $K=4$ to evaluate the key design components of XDecomposer under challenging multiphase conditions. We consider four structural ablations by removing multi-scale skip fusion, the geometry constraint, phase-guided modulation, or the global-context encoder $G(\cdot)$. Two alternative reconstruction strategies are also evaluated. Direct regression removes the mixture-aware masking constraint and directly predicts the component patterns, whereas hard-mask reconstruction binarizes the continuous assignment weights in overlapping regions. We further perform ablation experiments to quantify the effects of self-supervised pretraining, individual loss components, and loss weighting strategies. 

\begin{table}[h]
    \centering
    \caption{Ablation results on $K=4$.}
    \label{tab:ablation}
    \footnotesize
    \setlength{\tabcolsep}{3pt}
    \renewcommand{\arraystretch}{0.92}
    \begin{tabular}{@{}lcccc@{}}
        \toprule
        Variant &Pearson$\uparrow$ &Peak Shift$\downarrow$ & $\Delta$FWHM$\downarrow$ &Top-1$\uparrow$ \\
        \midrule
        Full XDecomposer & \textbf{0.8307} & \textbf{0.0486} & \textbf{5.2546} & \textbf{69.76} \\
        w/o multi-scale skip fusion & 0.4649 & 0.0635 & 17.7708 & 11.58 \\
        w/o geometry constraint & 0.4273 & 0.0709 & 11.9110 & 15.24 \\
        w/o phase-guided modulation & 0.4197 & 0.0744 & 12.2133 & 15.41 \\
        w/o global-context encoder $G(\cdot)$ & 0.4124 & 0.0813 & 13.5547 & 12.30 \\
        direct output & 0.3439 & 0.0682 & 16.3048 & 14.24 \\
        hard mask output & 0.1916 & 0.1276 & 22.3597 & 2.35 \\
        \bottomrule
    \end{tabular}
\end{table}

As shown in Table~\ref{tab:ablation}, the complete XDecomposer achieves the best performance across all decomposition and identification metrics. Removing multi-scale skip fusion produces the largest structural-ablation decrease in Pearson correlation, demonstrating the importance of retaining high-resolution peak information during reconstruction. Removing the geometry constraint, phase-guided modulation, or global-context encoder also substantially reduces Pearson and Top-1 accuracy, confirming that these modules provide complementary geometric, phase-specific, and long-range structural information. Direct output performs worse than all structural variants, while hard-mask reconstruction causes the most severe degradation. Furthermore, Table~\ref{tab:loss_ablation} shows that removing mixture consistency causes a moderate performance drop, while removing SI-SDR results in larger degradation in profile similarity and identification accuracy. The amplitude-only variant achieves 51.16\% Top-1 accuracy, confirming amplitude reconstruction as a strong base objective, while its lower performance than the full model demonstrates the value of auxiliary losses. Removing amplitude loss further decreases the Pearson correlation to 0.5982 and Top-1 accuracy to 47.24\%. Removing self-supervised pre-training causes the largest degradation, highlighting its critical role in robust decomposition and representation learning. These results demonstrate that continuous mixture-aware masking is essential for resolving overlapping diffraction signals. Further analysis is provided in Appendix~\ref{app:ablation}.

\section{Conclusion}
We present XDecomposer, a prior-free deep learning framework for whole-pattern decomposition in multiphase identification. By casting multiphase decomposition as a blind source separation problem, XDecomposer enables joint phase inference through set prediction and recovers component patterns without prior phase information. Experiments on simulated and real-world datasets show consistent advantages in pattern reconstruction, peak position and width fidelity, and phase identifiability. These results establish XDecomposer as a new perspective beyond traditional phase identification pipelines, avoiding the limitations of sequential retrieval-based methods and offering strong potential for scalable high-throughput applications.
\textbf{Limitations and Future Work.}
XDecomposer relies on fixed output slots with activity prediction, which works well for the 2--4 phase settings studied here but may limit flexibility for mixtures with more variable phase counts. The current mask-based reconstruction improves mixture consistency and physical plausibility, yet may be less robust under complex experimental distortions (e.g., peak broadening, shifts, and background variation), suggesting the need for more adaptive reconstruction strategies. 
By producing standardized one-dimensional PXRD patterns, XDecomposer enables seamless integration with existing identification (XQueryer \cite{xqueryer2025}), crystal structure prediction, refinement (WPEM \cite{cao2026ai}), and property prediction (PRDNet \cite{cao2025beyond}) models via lightweight preprocessing, including interpolation and normalization. These structural outputs support downstream applications such as property prediction, inverse materials design, autonomous discovery, and experimental validation.

\begin{ack}
This work was supported by the Strategic Priority Research Program of the Chinese Academy of Sciences (XDA0480102) and the National Natural Science Foundation of China (62576339).
\end{ack}

\medskip
\bibliographystyle{unsrtnat}
\bibliography{refs}

\newpage
\appendix

\section*{Appendix}
\addcontentsline{toc}{section}{Appendix}
\startcontents[appendix]
\printcontents[appendix]{}{1}{\setcounter{tocdepth}{2}}

\section{Methodology Details}
\label{app:method_appendix}

\subsection{Training Objectives}
\label{app:loss}

We present the optimization objectives for the pretraining and decomposition stages separately. The two objectives are not simply the same loss function applied to different data. Instead, they share several reconstruction preferences while serving different purposes: the former targets structural representation learning of PXRD patterns, whereas the latter addresses multi-phase decomposition, slot alignment, and active-phase detection.

\subsubsection{Pretraining Objective}
\label{app:loss_pretrain}

In Stage I, the model performs masked reconstruction pretraining on single-phase diffraction patterns. This objective is used to learn the parameters of the global context encoder $G(\cdot)$, while the reconstruction supervision is provided by an auxiliary decoder that is used only during pretraining. Given a single-phase pattern $x$ and its reconstruction $\hat{x}$, the pretraining loss is
\begin{equation}
    \mathcal{L}_{\mathrm{pre}}
    =
    \mathcal{L}_{\mathrm{amp}}^{\mathrm{pre}}
    +
    \lambda_{\mathrm{shape}}^{\mathrm{pre}}\mathcal{L}_{\mathrm{shape}}^{\mathrm{pre}}
    +
    \lambda_{\mathrm{geo}}^{\mathrm{pre}}\mathcal{L}_{\mathrm{geo}}^{\mathrm{pre}}.
    \label{eq:pretrain_loss}
\end{equation}
The amplitude term is defined as
\begin{equation}
    \mathcal{L}_{\mathrm{amp}}^{\mathrm{pre}}
    =
    \mathrm{mean}\!\left[
    |\hat{x}-x|\cdot(1+\alpha x)
    \right].
\end{equation}
This weighting increases supervision on major peak regions. The shape term is
\begin{equation}
    \mathcal{L}_{\mathrm{shape}}^{\mathrm{pre}}
    =
    -\mathrm{SI\text{-}SDR}(\hat{x},x).
    \label{sisdr}
\end{equation}
SI-SDR denotes the scale-invariant signal-to-distortion ratio. For a predicted pattern $\hat{\mathbf x}$ and reference pattern $\mathbf x$, it is defined as
$$
\alpha=\frac{\langle\hat{\mathbf x},\mathbf x\rangle}{\|\mathbf x\|_2^2},\qquad
\mathrm{SI\text{-}SDR}=10\log_{10}\frac{\|\alpha\mathbf x\|_2^2}{\|\hat{\mathbf x}-\alpha\mathbf x\|_2^2}.
$$
Equation\ref{sisdr} minimizes negative SI-SDR and therefore maximizes profile-shape agreement.

The geometry term operates on first- and second-order differences in the square-root domain:
\begin{equation}
    \mathcal{L}_{\mathrm{geo}}^{\mathrm{pre}}
    =
    \|\nabla \sqrt{\hat{x}}-\nabla \sqrt{x}\|_1
    +
    \beta\|\nabla^2 \sqrt{\hat{x}}-\nabla^2 \sqrt{x}\|_1.
\end{equation}
This stage is intended to capture local peak-shape statistics, long-range peak co-occurrence patterns, and transferable structural priors from single-phase data, rather than to directly learn multi-phase decomposition.

\subsubsection{Decomposition Objective}
\label{app:loss_sep}

In Stage II, the model transfers the pretrained global context encoder $G(\cdot)$ to the full decomposition model and keeps it frozen, while the decomposition-related modules are optimized on multiphase mixtures. Given the ground-truth single-phase contributions $\{y_k\}_{k=1}^{K}$ and the predicted contributions $\{\hat{y}_k\}_{k=1}^{K_{\max}}$, permutation-invariant training resolves the assignment ambiguity between output slots and target phases:
\begin{equation}
    \pi^{*}
    =
    \arg\min_{\pi}
    \sum_{k=1}^{K}\mathcal{L}_{\mathrm{decomp}}(\hat{y}_{\pi(k)},y_k).
    \label{eq:pit_appendix}
\end{equation}
The slot-wise decomposition loss is defined as
\begin{equation}
    \mathcal{L}_{\mathrm{decomp}}
    =
    \mathcal{L}_{\mathrm{amp}}
    +
    \lambda_{\mathrm{shape}}\mathcal{L}_{\mathrm{shape}}
    +
    \lambda_{\mathrm{geo}}\mathcal{L}_{\mathrm{geo}}.
    \label{eq:decomp_loss}
\end{equation}
The amplitude term is
\begin{equation}
    \mathcal{L}_{\mathrm{amp}}
    =
    \frac{1}{K}\sum_{k=1}^{K}
    \mathrm{mean}\!\left[
    |\hat{y}_{\pi^{*}(k)}-y_k|\cdot(1+\alpha y_k)
    \right].
\end{equation}
The shape term is defined as
\begin{equation}
    \mathcal{L}_{\mathrm{shape}}
    =
    -\mathrm{SI\text{-}SDR}(\hat{y},y).
\end{equation}

where SI-SDR denotes the scale-invariant signal-to-distortion ratio, is defined as
\begin{equation}
    \alpha=
    \frac{\langle\hat{y},y\rangle}
    {\|y\|_2^2},
    \qquad
    \mathrm{SI\text{-}SDR}
    =
    10\log_{10}
    \frac{\|\alpha y\|_2^2}
    {\|\hat{y}-\alpha y\|_2^2}.
\end{equation}

The geometry term is
\begin{equation}
    \mathcal{L}_{\mathrm{geo}}
    =
    \frac{1}{K}\sum_{k=1}^{K}
    \left(
    \|\nabla \hat{z}_{\pi^{*}(k)}-\nabla z_k\|_1
    +
    \beta\|\nabla^2 \hat{z}_{\pi^{*}(k)}-\nabla^2 z_k\|_1
    \right),
\end{equation}
where $z_k=\sqrt{y_k}$ and $\hat{z}_{\pi^{*}(k)}=\sqrt{\hat{y}_{\pi^{*}(k)}}$. These terms jointly constrain amplitude fidelity, spectral contour agreement, and local geometric consistency.

Based on the optimal assignment, the full decomposition objective is
\begin{equation}
    \mathcal{L}_{\mathrm{total}}
    =
    \mathcal{L}_{\mathrm{decomp}}^{\mathrm{PIT}}
    +
    \lambda_{\mathrm{act}}\mathcal{L}_{\mathrm{act}}
    +
    \lambda_{\mathrm{mix}}\mathcal{L}_{\mathrm{mix}}.
    \label{eq:total_loss_appendix}
\end{equation}
The activity supervision term is
\begin{equation}
    \mathcal{L}_{\mathrm{act}}
    =
    \frac{1}{K_{\max}}\sum_{k=1}^{K_{\max}}
    \mathrm{BCE}(p_k,c_k),
\end{equation}
where $c_k\in\{0,1\}$ indicates whether slot $k$ is matched to a true active phase under the optimal PIT assignment. Unmatched slots are treated as inactive and assigned label $0$. The mixture-consistency term is
\begin{equation}
    \mathcal{L}_{\mathrm{mix}}
    =
    \|\hat{x}-x\|_1,
    \qquad
    \hat{x}=\sum_{k=1}^{K_{\max}}\hat{y}_k.
\end{equation}

In summary, although the pretraining and decomposition stages share the same general reconstruction preferences: amplitude, shape, and geometry, their optimization goals differ substantially. The former learns transferable pure-phase structure representations, whereas the latter additionally incorporates permutation-invariant matching, activity supervision, and mixture consistency for multi-phase decomposition. 

\subsection{Implementation Details}
\label{app:impl}

All experiments were conducted on a single node with dual AMD EPYC 7742 CPUs and 8$\times$ NVIDIA GeForce RTX 3090 24GB GPUs, using Python 3.10 and PyTorch 2.5.1 with Distributed Data Parallel (DDP).Table~\ref{tab:training_efficiency} summarizes the training configurations and model efficiency statistics.

In the decomposition stage, the pretrained global context encoder $G(\cdot)$ is frozen. Accordingly, only 3.09M out of the 9.20M total parameters are trainable.

\begin{table}[htbp]
    \centering
    \caption{Training configurations and model efficiency summary.}
    \label{tab:training_efficiency}
    \setlength{\tabcolsep}{3pt}
    \renewcommand{\arraystretch}{0.8}
    \begin{tabular}{lcc}
        \toprule
        \textbf{Parameter} & \textbf{Pretraining} & \textbf{Decomposition} \\
        \midrule
        Backbone & \multicolumn{2}{c}{Transformer} \\
        Embedding dimension & \multicolumn{2}{c}{768} \\
        Attention heads & \multicolumn{2}{c}{12} \\
        Transformer layers & \multicolumn{2}{c}{4} \\
        Optimizer & \multicolumn{2}{c}{AdamW} \\
        Weight decay & \multicolumn{2}{c}{0.05} \\
        Batch size & \multicolumn{2}{c}{128} \\
        LR scheduler & \multicolumn{2}{c}{Cosine decay} \\
        Learning rate & $5\times10^{-4}$ & $2\times10^{-4}$ \\
        Epochs & 2000 & 1000 \\
        Warmup epochs & 50 & 20 \\
        \midrule
        Patch size / stride & 50 / 25 & -- \\
        Mask ratio & 70\% & -- \\
        Conv channel widths & -- & \texttt{\{48,96,192,384\}} \\
        Conv kernel sizes & -- & \texttt{\{15,8,8,10\}} \\
        Conv strides & -- & \texttt{\{1,2,2,5\}} \\
        Downsampling ratio & -- & $\times 20$ \\
        Max slots $K_{\max}$ & -- & 4 \\
        EMA at inference & -- & Enabled \\
        \midrule
        Total params (M) & 6.11 & 9.20 \\
        Trainable params (M) & 6.11 & 3.09 \\
        Est. GFLOPs & 1.2--2.0 & 2.5--3.5 \\
        CPU latency (ms) & $\sim$15.0 & $\sim$30.0 \\
        FP32 model size (MB) & $\sim$24 & $\sim$36 \\
        \bottomrule
    \end{tabular}
\end{table}

\subsection{From Structure to PXRD}
\label{xrd_sim}
The physical process by which a PXRD pattern is generated from a crystal is detailed below:

The diffraction vector  \(\mathbf{Q}\) is defined as 
\begin{equation}
    |\mathbf{Q}| = \frac{2\pi\sin{\theta}}{\lambda},						 
\end{equation}
where $\lambda$ is the wavelength of the scattered X-ray and $\theta=2\theta/2$, with $2\theta$ being the angle between the incident and diffracted beams. 
In three dimensions, scattering occurs only at a discrete set of reciprocal vectors, $\mathbf{K}$, forming the reciprocal lattice,
\begin{equation}
    \mathbf{K} = h\mathbf{a}^* + k\mathbf{b}^* + l\mathbf{c}^*,
\end{equation}
where $\mathbf{a}^*$, $\mathbf{b}^*$, and $\mathbf{c}^*$ are the reciprocal lattice vectors, and $h$, $k$, and $l$ are constants. 
The diffraction condition is defined as 
\begin{equation}
    \mathbf{Q}= \mathbf{K}.
\end{equation}

The diffraction intensity $I$ on each diffraction vector $\mathbf{Q}$ for a single phase is determined by 
\begin{equation}
    I(\mathbf{Q}) = SF^{*}F\varnothing LPOD + I^{\text{BG}},				 
\end{equation}
where $S$ denotes the scale factor, \(F\) is the structure factor, \(F^*\) is the complex conjugate of \(F\), $\varnothing$ is the profile function, $L$ is the Lorentz-polarization factor, $P$ is the multiplicity, $O$ is the preferred orientation factor, $D$ is the Debye-Waller factor, and $I^{\text{BG}}$ is the background intensity.

The structure factor is computed as:
\begin{equation}
    F =  \sum_{j=1}^N f_j e^{i \mathbf{Q} \cdot \mathbf{R}}, 
\end{equation}
where $f_j$ is the form factor~\cite{hubbell1974international} in XRD, $\mathbf{R}$ is the lattice coordinate of atom $j$, and $N$ is the total number of atoms in a lattice cell. 

The Lorentz-polarization $L$ is calculated by,
\begin{equation}
    L = \frac{1 + \cos^2{2\theta}}{\sin^2{\theta}\cos{\theta}},					 
\end{equation}
where $\theta=\arcsin{\left(\frac{\lambda|\mathbf{Q}|}{2\pi}\right)}$. The multiplicity $P$ is determined by counting the number of diffraction vectors present within an Ewald diffraction sphere. 
The Debye-Waller factor $D$ is calculated by,
\begin{equation}
    D = e^{-2M},								 
\end{equation}
where $M = \frac{6h^2 T}{m k \Theta^2}\left(\phi\left(\frac{\Theta}{T}\right) + \frac{\Theta}{4T}\right) \left(\sin^2{\theta}\right)/\lambda$, $h$ is Planck’s constant, $m$ is atom mass, $k$ is Boltzmann constant, $\Theta$ is the average characteristic temperature, $T$ is absolute temperature, and $\phi(\Theta/T)$ is the Debye function. 

The profile function ($\varnothing$) is a convolution of various factors, including diffraction effects, detector geometry, and noise contributions. The peaks arise from:

\begin{equation}
    y(x) = W \otimes G \otimes S.
\end{equation}

Here, \( \otimes\) denotes the convolution process, and \( W \), \( G \), and \( S \) represent the contributions to the observed XRD pattern from diffraction emission, instrumental factors, and the noise mixture, respectively. \( S \) is modeled as a Gaussian peak.

The \( W \) is a Voigt function~\cite{armstrong1967spectrum}, 
\begin{equation}
    W = \frac{1}{\sigma\sqrt{2\pi}} \int_{-\infty}^{\infty} \left[\frac{\gamma}{(2\theta-t)^2 + \gamma^2}\right]\exp\left(-\frac{(2\theta-t)^2}{2\sigma^2}\right)dt.
\end{equation}
Peak broadening is correlated with the Full Width at Half Maximum (FWHM, $\Gamma$), where $2\gamma = 2\sqrt{2\ln{2}}\sigma = \Gamma$~\cite{caglioti1958choice}. $\Gamma$ is calculated by Scherrer's equation~\cite{miranda2018limit} related to finite grain size.

The geometrical factor\cite{van1984peak}, \( G \), accounts for the actual dimensions of the detectors and the powder specimen. It is defined as follows:

\begin{equation}
G(2\alpha, 2\theta) = \frac{L}{4HSh \cos(2\alpha)} \int dz,
\end{equation}
where \( 2\alpha \) represents the Bragg angle. The detector is slit-shaped, with a height of \( 2H \), and the sample has a height of \( 2S \). \( L \) denotes the distance between the specimen and the detector. 

For multiphase systems, the observed diffraction intensity arises from the additive superposition of the scattering contributions from all constituent phases, which is a fundamental property of PXRD under the assumption that different phases scatter independently without significant interaction effects. The measured intensity can be expressed as

\begin{equation}
I(\mathbf{Q}_{\mathrm{mult}}) = \sum_{i=1}^{N} w_i \, I_i(\mathbf{Q}) + \sigma(\mathbf{Q}),
\end{equation}

where \(I_i(\mathbf{Q})\) denotes the diffraction intensity of the \(i\)-th phase at scattering vector \(\mathbf{Q}\), \(w_i\) represents its relative proportion (mixture coefficient) in the sample, and \(N\) is the total number of constituent phases. The term \(\sigma(\mathbf{Q})\) accounts for experimental noise, background signals, instrumental fluctuations, trace impurities, and minor unidentified components whose contributions are typically weak and lack clear physical guidance.

Each single-phase pattern \(I_i(\mathbf{Q})\) is influenced by multiple physical factors, including crystal structure, preferred orientation, crystallite size, strain broadening, peak overlap, and instrumental broadening etc. In multiphase systems, these effects accumulate through linear superposition, making the observed pattern substantially more sensitive to small variations in any individual phase. Even slight deviations in peak position, intensity, or line shape from one component phase can propagate to the final mixed pattern and affect phase identification and quantitative decomposition.

As a result, multiphase modeling is inherently more fragile than single-phase analysis: errors introduced at the single-phase level are amplified after superposition, while overlapping peaks further increase ambiguity in phase separation. This explains the significant sim-to-real performance gap observed in experimental datasets, where real measurements contain stronger structural variation and more complex noise sources than simulated data.

\section{Experimental Implementation}
\label{app:exp_appendix}

\subsection{Details of Ablation Study}
\label{app:ablation}

We conduct ablation studies on simulated four-phase mixtures with $K=4$ to evaluate the contribution of the main design components of XDecomposer under severe peak overlap and increased combinatorial ambiguity. This setting provides a challenging testbed for assessing spectral reconstruction quality, peak-structure preservation, and downstream identification accuracy.

We first examine the architectural and reconstruction components of XDecomposer. The evaluated variants include removing multi-scale skip fusion, removing the geometry constraint, removing phase-guided modulation, and removing the global-context encoder $G(\cdot)$. Two alternative reconstruction forms are also considered. In the \textit{direct regression output} variant, the mixture-aware masking pathway is replaced by direct component-pattern regression. In the \textit{hard mask output} variant, the continuous assignment weights are binarized to approximate discrete allocation in overlapping regions. The results are reported in Table~\ref{tab:ablation}.

The full XDecomposer model achieves the best performance across both decomposition and identification metrics, indicating that its advantage arises from the joint effect of its architectural modules and reconstruction strategy. All ablated variants exhibit clear degradation at $K=4$, confirming that each design choice is important for handling severe peak overlap. Among the structural variants, \textit{w/o multi-scale skip fusion} remains closest to the full model in most metrics; however, its $\Delta$FWHM increases sharply from $5.2546$ to $17.7708$. This substantial increase indicates that multi-scale skip connections are critical for preserving local peak details and accurate line-width information. Removing the \textit{geometry constraint} or \textit{phase-guided modulation} leads to consistent declines in spectral fidelity, peak-position accuracy, and retrieval performance, suggesting that the geometry constraint promotes physically plausible peak structures, while phase-guided modulation improves the separation of component-specific representations. The \textit{w/o global-context encoder $G(\cdot)$} variant performs worst among the structural ablations, underscoring the importance of global-context modeling for recovering coherent and identifiable single-phase patterns.

The comparison of reconstruction forms further demonstrates the importance of the output pathway. The \textit{direct regression output} variant underperforms the full model, indicating that direct component regression does not fully exploit the structural information contained in the input mixture. The \textit{hard mask output} variant yields the lowest overall performance among the reconstruction variants. This result shows that discretizing continuous assignments disrupts shared spectral structure in overlapping regions, impairing both reconstruction accuracy and downstream phase identification.

We further analyze the effects of self-supervised pre-training and the individual components of the composite learning objective. For the pre-training ablation, the self-supervised stage is removed and the model is trained directly from initialization using the same architecture and supervised training protocol. For the objective-component ablations, one component is removed at a time while all other settings are kept unchanged. These results are summarized in Table~\ref{tab:loss_ablation}.

\begin{table}[t]
\centering
    \caption{Ablation results for self-supervised pre-training and objective components on simulated four-phase mixtures with $K=4$.}
\label{tab:loss_ablation}
\begin{tabular}{lcccc}
\toprule
Variant & Pearson $\uparrow$ & Peak Shift ($^\circ$) $\downarrow$ & $\Delta$FWHM ($^\circ$) $\downarrow$ & Top-1 (\%) $\uparrow$ \\
\midrule
Full model                     & \textbf{0.8322} & \textbf{0.0480} & \textbf{5.2410}  & \textbf{69.68} \\
\quad w/o mixture-consistency loss & 0.7912 & 0.0524 & 5.4364  & 60.34 \\
\quad w/o SI-SDR loss              & 0.7612 & 0.0514 & 5.4952  & 58.35 \\
\quad amplitude loss only          & 0.6287 & 0.0573 & 5.6198  & 51.16 \\
\quad w/o amplitude loss           & 0.5982 & 0.0609 & 5.9110  & 47.24 \\
\quad w/o activity loss            & 0.4073 & 0.0777 & 13.1825 & 14.34 \\
\quad w/o pre-training             & 0.4067 & 0.0787 & 13.5071 & 14.00 \\
\bottomrule
\end{tabular}
\end{table}

The loss-component ablations show that each part of the learning objective contributes to the final performance. Removing the mixture-consistency term moderately degrades all metrics, reducing Top-1 accuracy from $69.68\%$ to $60.34\%$. Removing the SI-SDR term causes a larger reduction in profile similarity and identification accuracy, indicating that scale-invariant spectral matching is beneficial for recovering distinguishable component patterns. The \textit{amplitude loss only} variant still achieves $51.16\%$ Top-1 accuracy, showing that amplitude reconstruction provides a strong basic training signal. Nevertheless, its gap from the full model confirms that auxiliary constraints are necessary for accurate phase separation under heavy overlap. Removing the amplitude loss further reduces Pearson correlation to $0.5982$ and Top-1 accuracy to $47.24\%$, demonstrating that amplitude-level supervision remains indispensable for faithful profile recovery.

The most pronounced degradation occurs when activity supervision or self-supervised pre-training is removed. Without activity supervision, Top-1 accuracy drops to $14.34\%$ and $\Delta$FWHM increases to $13.1825^\circ$, indicating that explicit activity-related supervision is essential for associating recovered patterns with identifiable phases. Similarly, removing self-supervised pre-training reduces Top-1 accuracy to $14.00\%$ and produces the largest $\Delta$FWHM among these variants. This result suggests that pre-training provides a critical initialization for learning robust spectral representations before supervised decomposition, especially in the highly ambiguous $K=4$ setting.

\subsection{Large-scale Simulated Single-phase Generation}
\label{app:sim_gen}

The simulated dataset comprises a large-scale corpus of single-phase XRD patterns generated using PySimXRD \cite{bin2025simxrd} and subsequently mixed online during the training stage. Pattern generation was conducted on a high-performance computing cluster with SLURM job scheduling and a multi-process parallel framework, enabling the generation of multiple simulated patterns for each crystal ID under varied parameter configurations. This design improves coverage of diverse sample states and experimental conditions encountered in real-world settings. A fixed random seed is used to ensure reproducibility. The objective is not to produce idealized template patterns, but to introduce sufficient variation in peak positions, widths, backgrounds, and instrumental geometry, thereby enabling the model to learn transferable diffraction structure representations under realistic perturbations. Table~\ref{tab:sim_gen} summarizes the main parameter configurations used for simulated single-phase PXRD pattern.

\begin{table*}[htbp]
    \centering
    \caption{Main parameter configurations for simulated single-phase XRD pattern}
    \label{tab:sim_gen}
    \begin{tabular}{@{}lll@{}}
        \toprule
        Category & Parameter & Value / Range \\
        \midrule
        \multirow{6}{*}{Random perturbations} & Crystallite size (nm) & 10--120 \\
        & Thermal vibration coefficient & 0.01--0.2 \\
        & Zero shift & 0--0.2 \\
        & Detector--sample distance (mm) & 300--600 \\
        & Detector slit half-height (mm) & 3--8 \\
        & Sample half-height (mm) & 1--4 \\
        \midrule
        \multirow{4}{*}{Fixed parameters} & Preferred orientation factor & 0.15 \\
        & Background polynomial order & 6 \\
        & Mixture noise ratio & 0.02 \\
        & Lattice extinction / torsion ratio & 0.01 / 0.01 \\
        \midrule
        \multirow{3}{*}{Data control} & $2\theta$ scan range & 10$^\circ$--80$^\circ$ \\
        & Step size & 0.02$^\circ$ \\
        & Spectra per crystal ID & 20 \\
        \bottomrule
    \end{tabular}
\end{table*}

\subsection{RRUFF Experimental Dataset Construction and Preprocessing}
\label{app:rruff}

To establish a cross-domain zero-shot generalization benchmark, we constructed a unified dataset of 662 high-quality experimental PXRD patterns with corresponding crystal structure information from the publicly available RRUFF mineral database \cite{rruff2015}. The raw data consist of experimentally measured diffraction patterns and separately stored CIF structure files. The construction pipeline comprises four stages: raw data acquisition, structure--pattern matching with deduplication, WPEM \cite{cao2026ai} baseline correction, and database serialization with quality control.

\paragraph{Structure--pattern matching} Experimental patterns were matched to CIF files using mineral names as the primary key. When multiple PXRD samples existed for the same mineral species, only one representative sample was retained to ensure a unique structure--pattern pair. All CIF structures were subsequently parsed using \texttt{pymatgen}'s \texttt{SpacegroupAnalyzer} (\texttt{symprec = 0.1}) \cite{ong2013pymatgen}, and structures containing site disorder were strictly excluded, yielding 665 naturally ordered mineral structures.

\paragraph{Baseline correction}
Since experimentally measured PXRD patterns inherently contain background baselines (see Appendix~\ref{xrd_sim}), and these non-crystalline background signals can interfere with phase decomposition, we applied a unified automated baseline stripping procedure to all experimental patterns using the WPEM module \cite{cao2026ai}. This method combines frequency-domain decomposition, piecewise polynomial fitting, and Savitzky--Golay smoothing to estimate and subtract the background baseline. The resulting baseline-corrected net patterns were serialized as NumPy arrays in the database. Three samples were discarded due to insufficient data points or anomalous angular ranges, resulting in a final database of 662 valid records.

\paragraph{Database serialization}
During model training and evaluation, the data loader further applies sensitive nonlinear iterative peak (SNIP) clipping algorithms\cite{morhavc2008peak} interpolation to a standard angular grid of 3500 equidistant points over $2\theta \in [10^\circ, 80^\circ]$, followed by max-value normalization. This ensures comparability across samples on a unified input grid.

\subsection{Data Mixing Strategy}
\label{app:mixing}

During decomposition training, we construct training samples on the fly using an anchor-driven online mixing protocol, rather than relying on a pre-generated fixed dataset of multiphase mixtures. For each online sample, one single-phase pattern from the current data split is first selected as the anchor component, and the remaining components are then sampled from the same split. A simulated mixture is constructed from these components using randomly sampled mixing weights. Its physical basis follows the linear superposition approximation \cite{pecharsky2003fundamentals}, under which a multiphase diffraction pattern is approximated as the weighted sum of the diffraction contributions from its constituent phases.

The pre-prepared single-phase set covers 100,315 crystal samples, and each crystal sample is simulated under 20 different realistic conditions, yielding a total of \(100{,}315 \times 20\) single-phase PXRD patterns. For each crystal sample, we randomly select one pattern from the 20 candidates and use the resulting set as the basis for subsequent online mixing. Let the single-phase library corresponding to the current split be denoted by $\mathcal{D}^{(\mathrm{split})}=\{x^{(m)}\}_{m=1}^{M_{\mathrm{split}}}$, where $x^{(m)}\in\mathbb{R}_{+}^{L}$, $M_{\mathrm{split}}$ is the number of single-phase patterns in that split, and $L$ is the spectrum length on the common sampling grid. Online mixing is always performed after the train, validation, and test splits have been defined at the single-phase level. As a result, mixture samples for training, validation, and testing are generated only from the single-phase patterns within their respective splits, with no cross-split mixing of components.

For each online sample, the number of components is first drawn as
\begin{equation}
	N \sim \mathrm{Unif}\{2,3,4\}.
\end{equation}
An anchor single-phase pattern is then selected from $\mathcal{D}^{(\mathrm{split})}$, with index $a$, and the corresponding anchor component is written as $x_{\mathrm{anchor}}=x^{(a)}$. Given the anchor, the remaining $N-1$ single-phase patterns are sampled without replacement from $\mathcal{D}^{(\mathrm{split})}\setminus\{x^{(a)}\}$. Let their index set be $R=\{r_1,\dots,r_{N-1}\}\subset\{1,\dots,M_{\mathrm{split}}\}\setminus\{a\}$ with $|R|=N-1$. The resulting candidate component set is $\{x_i\}_{i=1}^{N}=\{x_{\mathrm{anchor}},x^{(r_1)},\dots,x^{(r_{N-1})}\}$.

The mixing weights are sampled from a Dirichlet distribution,
\begin{equation}
	w \sim \mathrm{Dir}(\alpha_1,\dots,\alpha_N),
\end{equation}
where $w=(w_1,\dots,w_N)$ satisfies $w_i\ge 0$ and $\sum_{i=1}^{N} w_i=1$. In the default setting, we use a symmetric Dirichlet prior, i.e., $\alpha_1=\cdots=\alpha_N=\alpha$, with $\alpha=1$. To prevent any component from becoming nearly unobservable after mixing, we further require
\begin{equation}
	w_i \ge 0.15, \qquad i=1,\dots,N.
\end{equation}
If a sampled weight vector does not satisfy this condition, it is discarded and resampled.

Given the sampled components and weights, the mixture pattern is constructed as
\begin{equation}
	x_{\mathrm{mix}}=\sum_{i=1}^{N} w_i x_i.
\end{equation}
We then add low-amplitude Gaussian noise to the mixture,
\begin{equation}
	\tilde{x}_{\mathrm{mix}} = x_{\mathrm{mix}} + \varepsilon, \qquad
	\varepsilon \sim \mathcal{N}(0,\sigma^2 I),
\end{equation}
where $\sigma$ is the preset noise level and $I$ is the identity matrix. To keep the model input and supervision targets on the same numerical scale, we apply max normalization with a shared scaling factor. Specifically, we define
\begin{equation}
	c = \max\!\left(
	\max_{1\le t\le L} \tilde{x}_{\mathrm{mix}}(t),\;
	\max_{1\le i\le N}\max_{1\le t\le L} x_i(t)
	\right),
\end{equation}
and obtain the normalized input and targets as
\begin{equation}
	\hat{x}_{\mathrm{mix}}=\frac{\tilde{x}_{\mathrm{mix}}}{c}, \qquad
	\hat{x}_i=\frac{x_i}{c}, \quad i=1,\dots,N.
\end{equation}
Each online sample is therefore represented as $(\hat{x}_{\mathrm{mix}},\{\hat{x}_i\}_{i=1}^{N},\{w_i\}_{i=1}^{N})$.

Three considerations motivated our adoption of the online mixing strategy. First, the space of multiphase mixtures grows rapidly with component choice, composition ratio, and noise perturbation, making it difficult for a fixed dataset to cover the resulting continuous combination space. Second, online mixing continuously generates new component combinations, weight configurations, and noise realizations during training, which increases sample diversity. Third, on-demand generation avoids the additional storage and data-management cost of pre-generating large numbers of mixture patterns, while allowing unified control over the component range, weight lower bound, and noise level.

\subsection{Data Split}
\label{app:split}

For the simulated data, the single-phase pattern library is partitioned at the crystal ID level rather than the pattern file level. This prevents patterns generated from the same crystal under different parameter perturbations from appearing simultaneously in the training, validation, and test sets, thereby eliminating potential data leakage. In other words, the split unit is the ``crystal instance,'' not the ``pattern sample,'' ensuring that the model does not merely memorize near-replicate patterns of the same structure under varying perturbations.

For the real-domain experiments, evaluation is uniformly based on the 662 RRUFF experimental single-phase patterns using 5-fold cross-validation. The dataset is partitioned into five disjoint subsets, with each subset serving as the test set in turn. The mean and standard deviation of all metrics are reported across the five folds.

\subsection{Baseline Implementation}
\label{app:baselines}

This section describes how the comparative baselines are instantiated within our experimental framework. Since the original objectives of these methods are not fully aligned, the comparison does not focus on whether they share exactly the same task definition as XDecomposer. Instead, we evaluate whether they can recover single-phase representations with physical interpretability and downstream identifiability under a unified evaluation protocol.

We divide the baselines into two categories: \textbf{domain-specific baselines} and \textbf{sequence baselines}. The former includes XQueryer, XRDAutoAnalyzer, and XRD\_Proportion\_Inference, all of which were originally developed for PXRD phase identification or quantitative analysis. The latter includes Transformer, iTransformer, and PatchTST, which are general-purpose sequence models adapted here to the PXRD decomposition setting.

\paragraph{XQueryer}
XQueryer is the first cross-chemical general model for single-phase powder XRD structure identification and is therefore used as a baseline in our study \cite{xqueryer2025}. Since the original XQueryer targets single-phase XRD identification, we adapt it into a multi-phase baseline by introducing a slot-based prediction head on mixture inputs. Specifically, instead of producing one identification result for a single pattern, the adapted model takes a mixed XRD pattern as input and predicts $S$ unordered phase slots; each slot contains a reconstructed component pattern, an estimated mixing ratio, and a classification logit over candidate phases. We train this variant with online-synthesized mixtures and use Hungarian matching to associate predicted slots with ground-truth phases. This modification preserves the main XQueryer backbone while enabling direct comparison under our unified multi-phase separation and identification benchmark.

\paragraph{XRDAutoAnalyzer}
XRDAutoAnalyzer is based on the automated XRD analysis framework of Szymanski \emph{et al.}~\cite{szymanski2021probabilistic,szymanski2024xrdpdf}, whose original goal is phase identification and refinement within a reference-driven analysis pipeline rather than blind multi-phase source separation. In our benchmark, we adapt it through an external wrapper instead of modifying its core model. Specifically, we first synthesize mixture patterns from multiple single-phase references, then feed each mixture to the official \texttt{SpectrumAnalyzer}, which returns candidate phases, confidence scores, refinement scalings, and reconstructed phase-wise spectra. We map these outputs to our protocol by aligning predicted phases to the ground truth, forming a fixed-size set of predicted component patterns, and deriving compatible activity signals for unified metric computation. This adaptation allows us to compare XRDAutoAnalyzer under the same separation, identification, and quantification metrics as other baselines, while making clear that it remains a retrieval-and-refinement pipeline rather than an end-to-end decomposition network. Owing to this design, XRDAutoAnalyzer is evaluated only in the simulated-domain setting and is not included in the main experimental-domain comparison.

\paragraph{XRD\_Proportion\_Inference}
XRD\_Proportion\_Inference is based on the deep-learning framework for XRD quantitative analysis introduced by Simonnet \emph{et al.}, together with its Vision Transformer extension~\cite{simonnet2024phasequant,simonnet2025vitxrd}. Since the original formulation is designed for proportion inference rather than source-resolved separation, we adapt it into a multi-phase benchmark baseline by changing the prediction target from global composition regression to component-pattern reconstruction. Concretely, instead of using the original output head that predicts phase proportions from a single mixed pattern, we employ a separation-oriented encoder--decoder variant that preserves the overall convolutional parameterization of the baseline while mapping an input mixture pattern to a fixed set of reconstructed component patterns. Training samples are generated by online mixing of multiple single-phase XRD references, with mixture cardinality and proportions sampled on the fly, and the model is supervised against the corresponding component-wise target patterns. The adapted outputs are then evaluated under our unified protocol for pattern reconstruction, active-phase detection, database retrieval, and quantitative estimation, allowing us to assess how well a proportion-oriented baseline can be repurposed for multi-phase decomposition.

\paragraph{Transformer}
The Transformer baseline uses a standard attention-based sequence-to-sequence architecture \cite{vaswani2017transformer}. It shares the same pretrained MAE encoder-side backbone as the other sequence baselines, including patch embedding, positional encoding, and the encoder. A Transformer decoder is then attached after the shared backbone to predict a fixed set of component patterns and their corresponding activity logits.

\paragraph{iTransformer}
The iTransformer baseline is built on the same pretrained MAE encoder backbone \cite{DBLP:journals/corr/abs-2310-06625}. Unlike the standard Transformer variant, it applies an inverted Transformer structure after the shared encoder and performs component-wise prediction in latent space. The model outputs a fixed set of reconstructed component patterns together with their activity logits.

\paragraph{PatchTST}
The PatchTST baseline also uses the same pretrained MAE encoder backbone and attaches an encoder-only PatchTST-style prediction head after the shared latent representation \cite{DBLP:conf/iclr/NieNSK23}. It directly predicts a fixed set of component patterns and the corresponding activity logits from the encoded latent features.

\subsection{Metric Definitions}
\label{app:metrics}

\paragraph{Pearson Correlation Coefficient}
Given a predicted pattern $\hat{y}$ and target pattern $y$, the Pearson correlation coefficient is defined as
\begin{equation}
    \rho(\hat{y},y)=
    \frac{\sum_i (\hat{y}_i-\bar{\hat{y}})(y_i-\bar{y})}
    {\sqrt{\sum_i (\hat{y}_i-\bar{\hat{y}})^2}\sqrt{\sum_i (y_i-\bar{y})^2}}.
\end{equation}
This metric measures overall pattern shape consistency but does not explicitly distinguish peak shift from peak width distortion.

\paragraph{Mean Peak-Position Deviation}
Let the positions of the $m$-th matched peak pair be $\hat{\theta}_m$ and $\theta_m$. The mean peak-position deviation is defined as
\begin{equation}
    \overline{\Delta 2\theta} = \frac{1}{M}\sum_{m=1}^{M}|\hat{\theta}_m-\theta_m|.
\end{equation}
This metric directly reflects local peak geometry fidelity.

\paragraph{FWHM Error}
Let the full widths at half maximum of a matched peak pair be $\widehat{\mathrm{FWHM}}_m$ and $\mathrm{FWHM}_m$. The FWHM error is
\begin{equation}
    \Delta \mathrm{FWHM} = \frac{1}{M}\sum_{m=1}^{M}|\widehat{\mathrm{FWHM}}_m-\mathrm{FWHM}_m|.
\end{equation}
This metric reflects peak width recovery and is directly related to crystallographic line shape fidelity.

\paragraph{Top-1 / Top-10 Retrieval Accuracy}
Let $\hat{s}_n$ be a predicted single-phase pattern and $r_j$ a reference pattern in the database. An initial similarity function is first employed for coarse candidate recall; we use cosine similarity as the first-stage retrieval score:
\begin{equation}
    \mathrm{sim}_{\mathrm{base}}(\hat{s}_n,r_j) = \frac{\hat{s}_n^\top r_j}{\|\hat{s}_n\|_2 \, \|r_j\|_2}.
\end{equation}
Based on this score, the top $M$ candidates are selected to form a candidate set $\mathcal{C}_M(\hat{s}_n)$. A reranking module then recomputes scores for entries in the candidate set:
\begin{equation}
    \mathrm{sim}_{\mathrm{rerank}}(\hat{s}_n,r_j) = f_{\mathrm{rerank}}(\hat{s}_n,r_j), \quad r_j \in \mathcal{C}_M(\hat{s}_n),
\end{equation}
and the final retrieval list is obtained by reordering according to $\mathrm{sim}_{\mathrm{rerank}}$.

\subsection{Robustness Analysis}
We evaluate the robustness of our method in two challenging scenarios: resolving low-content phases and separating highly overlapping component patterns.

\begin{table}[h]
    \centering
    \caption{Effect of the minimum phase proportion.}
    \label{tab:minor_phase}
    \renewcommand{\arraystretch}{0.92}
    \begin{tabular*}{0.8\columnwidth}{@{\extracolsep{\fill}}cccccc@{}}
        \toprule
        $K$ & Min.$w_k$ &
        Pearson $\uparrow$ &
        \makecell{Peak Shift $\downarrow$} &
        $\Delta$FWHM $\downarrow$ &
        \makecell{Top-1 (\%) $\uparrow$} \\
        \midrule
        \multirow{3}{*}{2}
        & 0.15 & 0.9691 & 0.0272 & 2.4330 & 87.92 \\
        & 0.10 & 0.9644 & 0.0260 & 2.3560 & 87.79 \\
        & 0.05 & 0.9553 & 0.0245 & 2.2186 & 85.89 \\
        \midrule
        \multirow{3}{*}{3}
        & 0.15 & 0.9101 & 0.0389 & 3.7836 & 81.13 \\
        & 0.10 & 0.8971 & 0.0369 & 3.5879 & 78.69 \\
        & 0.05 & 0.8727 & 0.0343 & 3.3141 & 75.11 \\
        \midrule
        \multirow{3}{*}{4}
        & 0.15 & 0.8322 & 0.0480 & 5.2410 & 69.68 \\
        & 0.10 & 0.8107 & 0.0457 & 4.8147 & 66.19 \\
        & 0.05 & 0.7755 & 0.0422 & 4.3668 & 61.44 \\
        \bottomrule
    \end{tabular*}
\end{table}

Table~\ref{tab:minor_phase} reports the minimum phase proportions considered, ranging from 0.15 to 0.05. As the proportion of minor phases decreases, Pearson correlation drops by 1.4\%, 4.1\%, and 6.8\% for $K=2,3,4$, respectively, while Top-1 accuracy decreases by 2.03, 6.02, and 8.24 percentage points. The larger degradation at higher $K$ arises from the increased difficulty in resolving weak reflections, which are progressively masked by dominant and overlapping peaks from other components. This behavior is physically expected, as low-content phases produce weaker diffraction signals that are more susceptible to interference from dominant phases, background contributions, and experimental uncertainties.

For the similarity-controlled benchmark, additional components are selected from the farthest, intermediate, or nearest 10\% of test phases ranked by cosine similarity. Table~\ref{tab:pattern_similarity} shows that the medium condition closely matches the original random-mixture benchmark. From medium to high similarity, Pearson decreases from 0.9097 to 0.8739 for $K=3$ and from 0.8275 to 0.7945 for $K=4$, while Top-1 decreases by 4.01 and 5.95 percentage points. For $K=2$, Top-1 accuracy remains nearly unchanged, although reconstruction and peak-based metrics show noticeable degradation. These results indicate that both low phase proportions and high pattern similarity become increasingly challenging as the number of constituent phases grows.

\begin{table}[h]
    \centering
    \caption{Effect of inter-phase pattern similarity.}
    \label{tab:pattern_similarity}
    \renewcommand{\arraystretch}{0.8}
    \begin{tabular*}{0.8\columnwidth}{@{\extracolsep{\fill}}clccccc@{}}
        \toprule
        $K$ & Sim. &
        Cosine &
        Pearson $\uparrow$ &
        \makecell{Peak Shift $\downarrow$} &
        $\Delta$FWHM $\downarrow$ &
        \makecell{Top-1 (\%) $\uparrow$} \\
        \midrule
        \multirow{3}{*}{2}
        & Low  & 0.1439 & 0.9839 & 0.0222 & 1.5880 & 85.94 \\
        & Med. & 0.2866 & 0.9692 & 0.0272 & 2.4665 & 88.43 \\
        & High & 0.5208 & 0.9542 & 0.0301 & 2.9770 & 88.39 \\
        \midrule
        \multirow{3}{*}{3}
        & Low  & 0.1508 & 0.9554 & 0.0298 & 2.2315 & 81.95 \\
        & Med. & 0.2942 & 0.9097 & 0.0395 & 3.8291 & 81.34 \\
        & High & 0.5221 & 0.8739 & 0.0441 & 4.7655 & 77.33 \\
        \midrule
        \multirow{3}{*}{4}
        & Low  & 0.1535 & 0.9065 & 0.0358 & 2.7895 & 75.42 \\
        & Med. & 0.3000 & 0.8275 & 0.0494 & 5.2740 & 69.40 \\
        & High & 0.5233 & 0.7945 & 0.0550 & 6.5606 & 63.45 \\
        \bottomrule
    \end{tabular*}
\end{table}

Additional results demonstrate that the model effectively suppresses spurious peaks in inactive slots (Appendix \ref{app:inactive_slot_analysis}), achieves the best decomposition and identification performance when using the full $2\theta$ range (Appendix \ref{app:angular_range_sensitivity}), and remains robust to threshold selection (Appendix \ref{app:activity_threshold_sensitivity}).

\subsection{Additional Robustness Analysis}
\label{app:additional_diagnostic_analysis}

This section provides additional sensitivity analyses concerning inactive output slots, the angular range of the input pattern, and the activity
threshold used to select predicted phases.

\subsubsection{Profile-Factor Evaluation}
\label{app:profile_factor_evaluation}

To further quantify the agreement between the predicted component profiles and their corresponding ground-truth patterns, we report the conventional profile
factor $R_p$ and the weighted profile factor $R_{wp}$. For a matched predicted
profile $\hat{y}$ and ground-truth profile $y$, sampled at $N$ diffraction-angle
positions, these metrics are defined as
\begin{equation}
    R_p=\frac{\sum_{i=1}^{N}\left|y_i-\hat{y}_i\right|
    }{\sum_{i=1}^{N}\left|y_i\right|}
    \times 100\%,\qquad 
    R_{wp}
    =\sqrt{\frac{\sum_{i=1}^{N}w_i\left(y_i-\hat{y}_i\right)^2}{\sum_{i=1}^{N}w_i y_i^2}}
    \times 100\%,
    \label{eq:profile_factor_rwp}
\end{equation}
where $y_i$ and $\hat{y}_i$ denote the ground-truth and predicted intensities,
respectively, at the $i$-th diffraction-angle position, and $w_i$ denotes the
corresponding weight.
While $R_p$ measures the normalized absolute profile discrepancy, $R_{wp}$
places greater emphasis on larger intensity deviations through its
squared-error formulation.

\begin{table}[htbp]
    \centering
    \caption{Profile factors for mixtures containing different numbers of phases.}
    \label{tab:profile_factors}
    \setlength{\tabcolsep}{12pt}
    \begin{tabular}{c cc}
        \toprule
        $K$
        & $R_p$ (\%) $\downarrow$
        & $R_{wp}^{(w=1)}$ (\%) $\downarrow$ \\
        \midrule
        2 & 13.20 & 10.17 \\
        3 & 27.74 & 23.60 \\
        4 & 45.04 & 38.89 \\
        \bottomrule
    \end{tabular}
\end{table}

As $K$ increases, both profile factors increase, reflecting the greater
difficulty of decomposing mixtures containing a larger number of strongly
overlapping diffraction patterns. Nevertheless, the profile-factor results are
consistent with the performance trends observed using Pearson correlation,
peak-position error, and peak-width error.

\subsubsection{Behavior of Inactive Output Slots}
\label{app:inactive_slot_analysis}

We first examine whether slots classified as inactive retain peak-like signals and whether such signals should be interpreted as additional crystalline phases.
After permutation-invariant matching, each output slot is associated either with
a real phase target or with a zero-padding target. The latter corresponds to an
extra output slot that should ideally be suppressed, whereas a real-phase target
classified as inactive represents a missed phase.

To quantify the residual structure in these outputs, peaks are detected using a normalized height threshold of $h>0.05$ and a non-maximum suppression (NMS)
window of 10 samples. Table~\ref{tab:inactive_slot_statistics} reports the
inactive rate, median relative energy, and peak-count statistics for each
PIT-matched target category.
\begin{table}[h]
    \centering
    \caption{
        Statistics of output slots classified as inactive after PIT matching. Padding denotes an extra slot matched to a zero target, whereas a real
        phase denotes a true component missed by the activity prediction.
        Peaks are detected using $h>0.05$ and an NMS window of 10 samples.}
    \label{tab:inactive_slot_statistics}
    \setlength{\tabcolsep}{5pt}
    \small
        \begin{tabular}{c l c c c c}
            \toprule
            $K$
            & PIT-matched target& Inactive rate (\%)& Median relative energy
            & No peak (\%)& $\geq 2$ peaks (\%) \\
            \midrule
            \multirow[c]{2}{*}{2}
            & Padding (extra slot)& 96.70 & 0.0011 & 94.50 & 3.29 \\
            & Real phase (missed)& 1.66 & 0.0246 & 54.35 & 33.93 \\
            \midrule
            \multirow[c]{2}{*}{3}
            & Padding (extra slot)& 73.35 & 0.0210 & 67.11 & 18.31 \\
            & Real phase (missed)& 5.95 & 0.0329 & 45.09 & 39.56 \\
            \bottomrule
        \end{tabular}%
\end{table}

Most padding-matched inactive outputs are strongly suppressed. For $K=2$, $94.50\%$ of these outputs contain no detectable peaks, and their median
relative energy is only $0.0011$. The $K=3$ setting is more challenging and
retains more weak peak fragments; nevertheless, $67.11\%$ of the corresponding
outputs remain peak-free, and the median relative energy is still low at
$0.0210$. In contrast, inactive outputs matched to real phases exhibit higher
relative energy and substantially more multi-peak structure, indicating that
they correspond to activity-detection false negatives rather than hallucinated
additional phases.

This behavior is also constrained by the mask-based output parameterization:
\begin{equation}
    m_k=\sigma(U_k),
    \qquad
    \hat{y}_k=m_k\odot x,
    \qquad
    0\leq \hat{y}_k\leq x.
    \label{eq:inactive_slot_constraint}
\end{equation}
Each predicted contribution is therefore a pointwise attenuated component of
the observed mixture and cannot introduce intensity beyond the input pattern.
Weak structures retained by inactive slots must originate from intensity
already present in the mixture. They are thus most plausibly attributed to
unresolved residual intensity, background or noise retained in the input, or
duplicated allocation of reflections from existing phases, rather than to
independently generated crystalline phases.

\subsubsection{Sensitivity to the Input Angular Range}
\label{app:angular_range_sensitivity}

We next evaluate the effect of the available diffraction-angle range. In
addition to the complete $2\theta$ interval of $10^\circ$--$80^\circ$, we
consider four representative restricted windows: $10^\circ$--$20^\circ$, $30^\circ$--$40^\circ$, $50^\circ$--$60^\circ$, and $70^\circ$--$80^\circ$. The results are reported in Table~\ref{tab:angular_range_sensitivity}.
\begin{table}[htbp]
    \centering
    \caption{
        Decomposition and phase-identification performance under different $2\theta$ input ranges. Bold values denote the best result for each
        mixture cardinality $K$.}
    \label{tab:angular_range_sensitivity}
    \setlength{\tabcolsep}{4pt}
    \small
        \begin{tabular}{c c c c c c c}
            \toprule
            $K$
            & $2\theta$ range& Pearson $\uparrow$& Peak shift ($^\circ$) $\downarrow$& $\Delta$FWHM ($^\circ$) $\downarrow$& Top-1 (\%) $\uparrow$& Top-10 (\%) $\uparrow$ \\
            \midrule
            \multirow[c]{5}{*}{2}
            & Full ($10^\circ$--$80^\circ$)
            & \textbf{0.9691}& \textbf{0.0272}& \textbf{2.4330}& \textbf{87.92}
            & \textbf{92.50} \\
            & $10^\circ$--$20^\circ$
            & 0.6503 & 0.0664 & 7.3907 & 12.10 & 20.17 \\
            & $30^\circ$--$40^\circ$
            & 0.7485 & 0.0383 & 5.7770 & 30.73 & 41.27 \\
            & $50^\circ$--$60^\circ$
            & 0.7143 & 0.0389 & 6.5156 & 39.40 & 51.52 \\
            & $70^\circ$--$80^\circ$
            & 0.6929 & 0.0433 & 6.0569 & 45.83 & 59.59 \\

            \midrule

            \multirow[c]{5}{*}{3}
            & Full ($10^\circ$--$80^\circ$)
            & \textbf{0.9101}& \textbf{0.0389}& \textbf{3.7836}& \textbf{81.13}
            & \textbf{86.81} \\
            & $10^\circ$--$20^\circ$
            & 0.5505 & 0.0716 & 9.9861 & 9.29 & 15.10 \\
            & $30^\circ$--$40^\circ$
            & 0.6084 & 0.0454 & 7.2331 & 16.91 & 23.66 \\
            & $50^\circ$--$60^\circ$
            & 0.5637 & 0.0468 & 8.3543 & 22.83 & 30.58 \\
            & $70^\circ$--$80^\circ$
            & 0.5465 & 0.0527 & 8.3102 & 27.74 & 37.13 \\

            \midrule

            \multirow[c]{5}{*}{4}
            & Full ($10^\circ$--$80^\circ$)
            & \textbf{0.8322}& \textbf{0.0480}& \textbf{5.2410}& \textbf{69.68}
            & \textbf{76.21} \\
            & $10^\circ$--$20^\circ$
            & 0.4903 & 0.0713 & 11.3767 & 7.22 & 11.63 \\
            & $30^\circ$--$40^\circ$
            & 0.5043 & 0.0506 & 9.0105 & 8.87 & 13.00 \\
            & $50^\circ$--$60^\circ$
            & 0.4671 & 0.0524 & 9.5928 & 12.53 & 17.90 \\
            & $70^\circ$--$80^\circ$
            & 0.4516 & 0.0567 & 9.7295 & 16.12 & 22.23 \\
            \bottomrule
        \end{tabular}%
\end{table}

For $K=2$--$4$, the complete $10^\circ$--$80^\circ$ input consistently
achieves the best performance across all decomposition and identification metrics. This result demonstrates that the angular regions contain complementary information that cannot be recovered from any isolated window.

Among the restricted ranges, $30^\circ$--$40^\circ$ consistently provides the highest Pearson correlation and the lowest peak-position and peak-width errors,
indicating the best preservation of the overall pattern profile and local peak geometry. By contrast, $70^\circ$--$80^\circ$ consistently produces the highest Top-1 and Top-10 retrieval accuracy among the restricted ranges. The latter observation may be attributed to the greater angular sensitivity of high-angle reflections to small differences in lattice spacing, despite the smaller number and generally lower intensity of reflections in this region. Because no restricted window is optimal for both reconstruction and retrieval,
the complete diffraction range is required for robust multiphase analysis.

\subsubsection{Sensitivity to the Activity Threshold}
\label{app:activity_threshold_sensitivity}

The activity probability $p_k$ is used to select the final set of predicted phases:
\begin{equation}
    \hat{\mathcal{Y}}^{*}
    =
    \left\{
        \hat{y}_k \,\middle|\, p_k>\tau
    \right\},
    \label{eq:activity_threshold_selection}
\end{equation}
where $\tau$ denotes the activity threshold. To evaluate sensitivity to this
parameter, we vary $\tau$ from $0$ to $0.8$ and report the corresponding
performance in Table~\ref{tab:activity_threshold_sensitivity}.

\begin{table}[htbp]
    \centering
    \caption{
        Sensitivity to the activity threshold $\tau$. Bold values indicate the setting used in the fixed-$K$ experiments rather than the numerical
        optimum of each individual column.
    }
    \label{tab:activity_threshold_sensitivity}
    \setlength{\tabcolsep}{4pt}
    \small
        \begin{tabular}{c c c c c c c}
            \toprule
            $K$
            & $\tau$& Pearson $\uparrow$& Peak shift ($^\circ$) $\downarrow$
            & $\Delta$FWHM ($^\circ$) $\downarrow$& Top-1 (\%) $\uparrow$
            & Top-10 (\%) $\uparrow$ \\
            \midrule

            \multirow[c]{6}{*}{2}
            & \textbf{0.0}& \textbf{0.9691}& \textbf{0.0272}& \textbf{2.4330}
            & \textbf{87.92}& \textbf{92.50} \\
            & 0.2 & 0.9668 & 0.0274 & 2.4673 & 88.10 & 92.28 \\
            & 0.4 & 0.9630 & 0.0274 & 2.4687 & 87.90 & 92.02 \\
            & 0.5 & 0.9609 & 0.0274 & 2.4670 & 87.74 & 91.83 \\
            & 0.6 & 0.9580 & 0.0273 & 2.4645 & 87.51 & 91.58 \\
            & 0.8 & 0.9477 & 0.0272 & 2.4536 & 86.69 & 90.63 \\

            \midrule

            \multirow[c]{6}{*}{3}
            & \textbf{0.0}& \textbf{0.9101}& \textbf{0.0389}& \textbf{3.7836}
            & \textbf{81.13}& \textbf{86.81} \\
            & 0.2 & 0.8959 & 0.0389 & 3.7997 & 80.57 & 86.17 \\
            & 0.4 & 0.8800 & 0.0389 & 3.8013 & 79.71 & 85.05 \\
            & 0.5 & 0.8710 & 0.0389 & 3.8015 & 79.14 & 84.37 \\
            & 0.6 & 0.8598 & 0.0388 & 3.8008 & 78.36 & 83.39 \\
            & 0.8 & 0.8345 & 0.0387 & 3.7834 & 76.31 & 81.06 \\

            \midrule

            \multirow[c]{6}{*}{4}
            & \textbf{0.0}& \textbf{0.8322}& \textbf{0.0480}& \textbf{5.2410}
            & \textbf{69.68}& \textbf{76.21} \\
            & 0.2 & 0.8272 & 0.0484 & 5.2352 & 69.70 & 76.07 \\
            & 0.4 & 0.8206 & 0.0484 & 5.2358 & 69.45 & 75.73 \\
            & 0.5 & 0.8150 & 0.0484 & 5.2367 & 69.20 & 75.42 \\
            & 0.6 & 0.8075 & 0.0483 & 5.2329 & 68.88 & 75.02 \\
            & 0.8 & 0.7794 & 0.0481 & 5.2056 & 67.53 & 73.38 \\

            \bottomrule
        \end{tabular}%
\end{table}

In the controlled fixed-$K$ experiments, we use $\tau=0$ so that no candidate component is discarded by activity thresholding. This protocol prevents
phase-number selection errors from being conflated with the intrinsic
decomposition performance and retains weak but valid phases under severe peak
overlap and noise.

As $\tau$ increases, Pearson correlation decreases because an increasing number of low-confidence but valid components are removed. Retrieval accuracy remains approximately stable at low thresholds and subsequently decreases, whereas the
peak-shift and FWHM errors vary only marginally. This indicates that thresholding
primarily affects phase retention rather than the local peak geometry of the
components that remain selected.

When the true number of phases is known, returning the top-$K$ slots ranked by $p_k$ avoids the need to select a global threshold. When $K$ is unknown, the
range $\tau\in[0,0.2]$ provides a practical operating region that suppresses
low-confidence outputs while retaining most valid phases. For deployment under
substantial domain shift, the threshold should be calibrated using a
representative validation set.

\subsection{Dataset Analysis}
\label{app:data_analysis}

Table~\ref{tab:data_summary} summarizes the basic statistics of the simulated dataset \textit{UniqCryLabeled} and the experimental dataset \textit{UniqRruffCrystal}. The experimental subset contains only 662 samples, whereas the simulated library contains 100,315 structures. The experimental subset also has narrower elemental coverage, with 55 unique elements compared with 84 in the simulated library. In addition, the experimental subset has lower median density (3.60 vs.\ 4.44~g/cm$^3$), smaller median volume per atom (12.3 vs.\ 15.1~\AA$^3$), and a larger median number of atoms per unit cell (36 vs.\ 20). These statistics indicate that the experimental set is not a uniform sample of the simulated structure space, but exhibits clear biases in both chemical composition and structural complexity.

\begin{table}[htbp]
    \centering
    \caption{Statistical comparison of the simulated and experimental datasets.}
    \label{tab:data_summary}
    \setlength{\tabcolsep}{6pt}
    \renewcommand{\arraystretch}{0.9}
    \begin{tabular}{lcc}
        \toprule
        Metric & \textit{UniqCryLabeled} & \textit{UniqRruffCrystal} \\
        \midrule
        Total entries & 100,315 & 662 \\
        Unique elements & 84 & 55 \\
        Median density (g/cm$^3$) & 4.44 & 3.60 \\
        Median Vol/Atom (\AA$^3$) & 15.1 & 12.3 \\
        Median $N_{\mathrm{atoms}}$ & 20 & 36 \\
        Dominant crystal system & Monoclinic & Monoclinic \\
        Most frequent space group & 1 & 14 \\
        \bottomrule
    \end{tabular}
\end{table}

\begin{figure*}[htbp]
    \centering
    \includegraphics[width=1\linewidth]{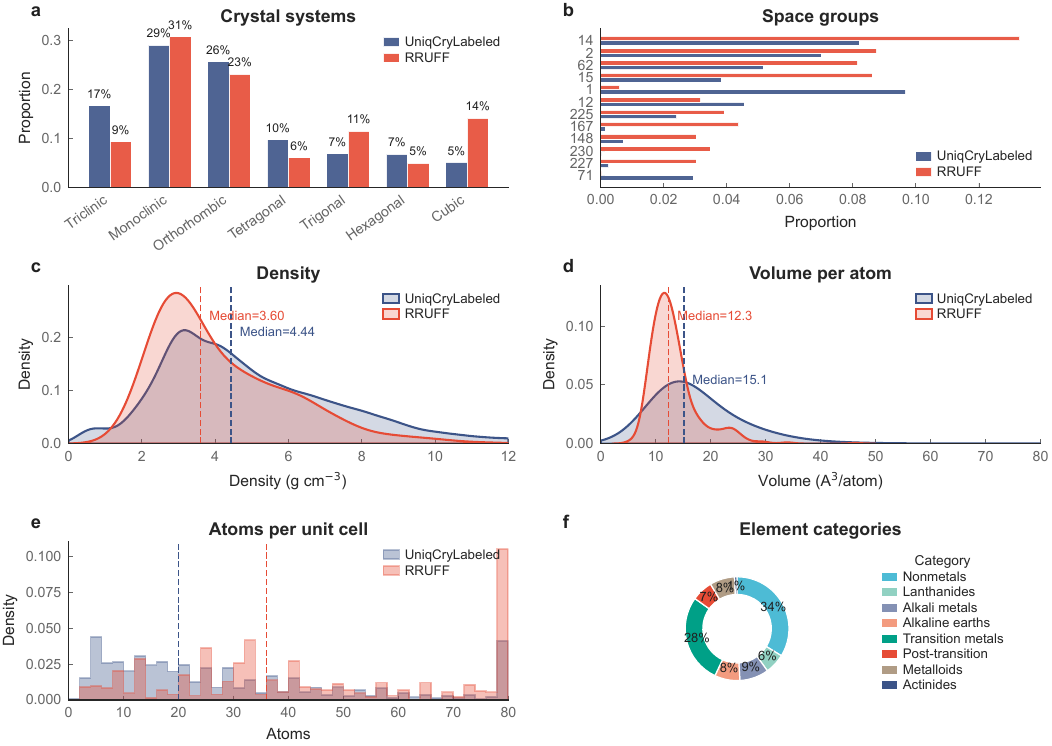}
    \caption{Comparison of categorical and physical attribute distributions between simulated and experimental datasets. Top: crystal system and high-frequency space group distributions; Bottom: density, volume per atom, and number of atoms per unit cell. The experimental subset exhibits systematic shifts in both symmetry composition and structural scales.}
    \label{fig:data_categorical_physical}
\end{figure*}

Figure~\ref{fig:data_categorical_physical} further shows systematic differences in categorical and continuous attributes. Both datasets are dominated by monoclinic structures, but the experimental subset has higher proportions of cubic and trigonal systems and lower proportions of triclinic, tetragonal, and hexagonal systems. The high-frequency space-group distribution also changes substantially: space group 1 is the most common in the simulated library, whereas space group 14 dominates the experimental subset. The continuous attributes follow the same trend. The experimental subset is shifted toward lower densities, smaller volumes per atom, and larger numbers of atoms per unit cell. These results indicate that the two domains differ not only in sample scale, but also in symmetry composition and local structural scale.

The chemical distributions show a similar pattern. The simulated library spans a broader elemental range, whereas the experimental subset is concentrated in typical mineral chemistries. O and Si are frequent in both datasets, but the experimental subset is relatively enriched in common mineral-forming elements such as Ca, Mg, H, P, Na, and Fe. This indicates that the experimental domain occupies a more localized region of chemical space rather than broadly covering the simulated compositional distribution.

\begin{figure*}[htbp]
    \centering
    \includegraphics[width=1\linewidth]{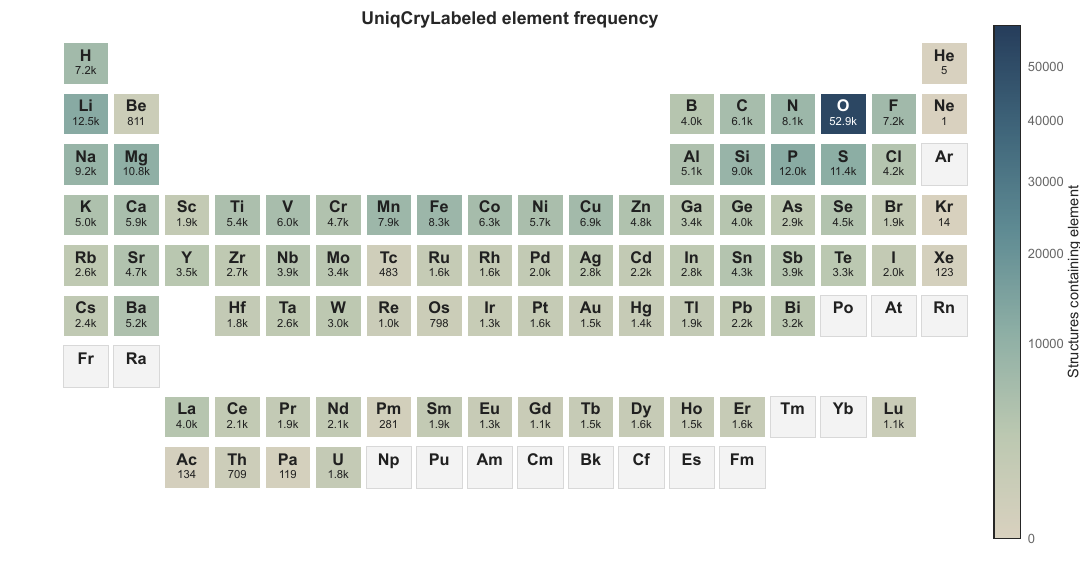}
 
    \includegraphics[width=1\linewidth]{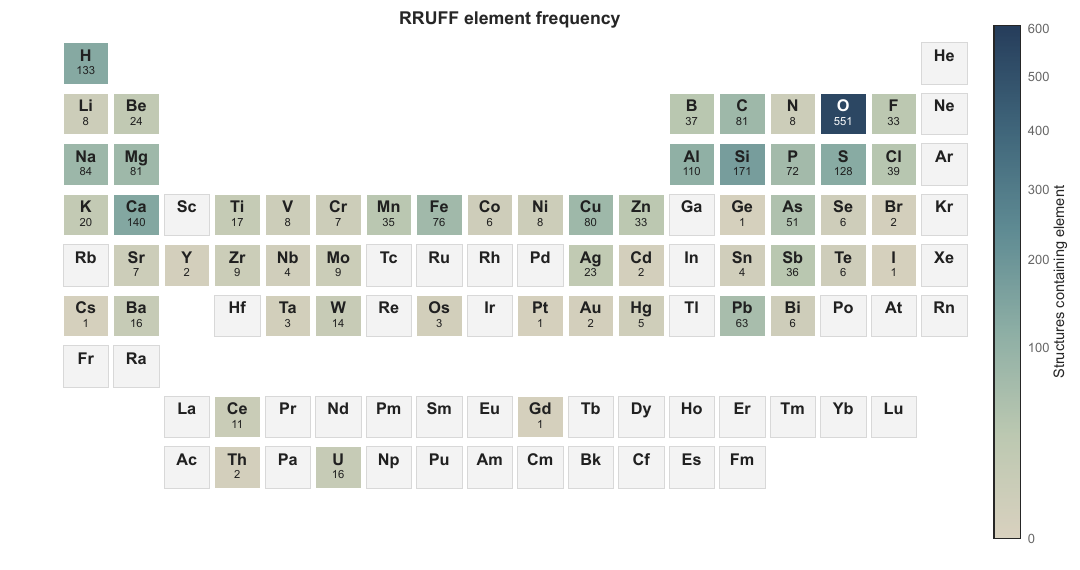}
    \caption{Element frequency in the simulated library (top) and the experimental RRUFF subset (bottom). The experimental subset covers a narrower chemical range and is concentrated around common mineral-forming elements.}
    \label{fig:data_chemistry}
\end{figure*}

\begin{figure*}[htbp]
    \centering
    \includegraphics[width=1\linewidth]{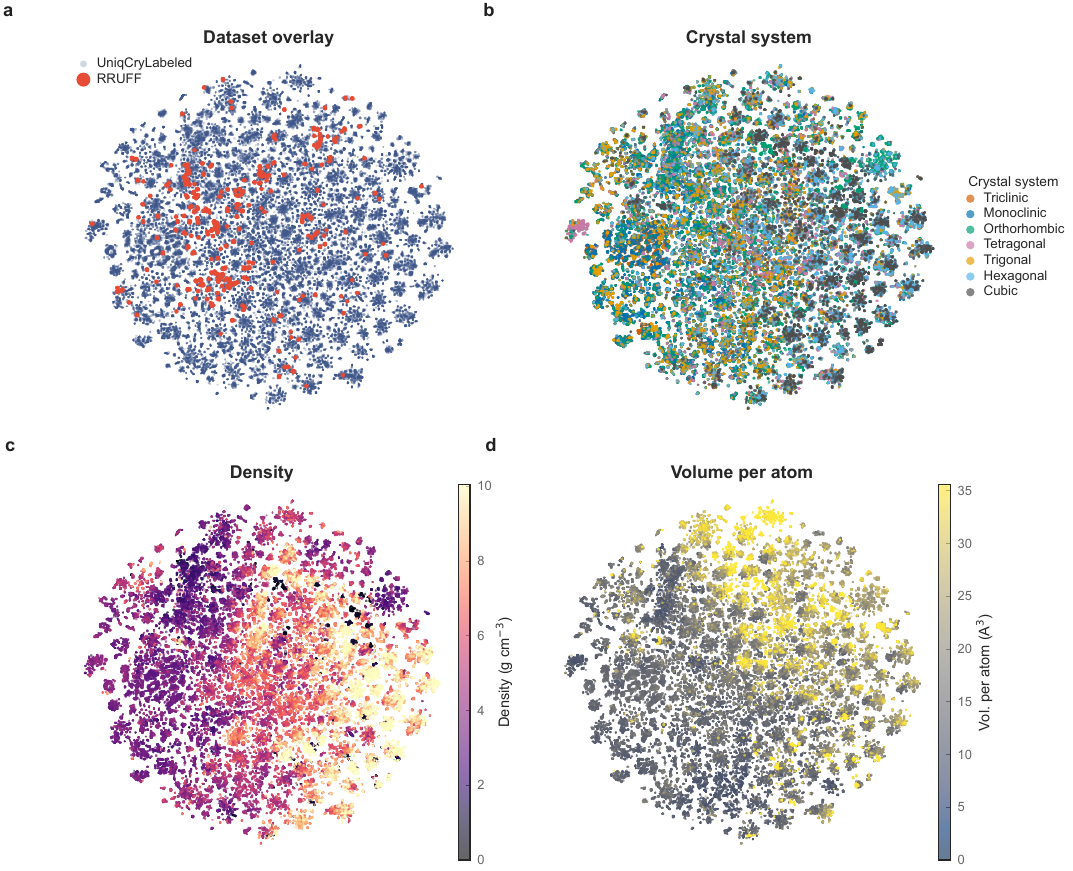}
    \caption{Joint manifold visualization of the simulated and experimental datasets. The experimental subset overlaps with the global simulated manifold but concentrates in several localized regions rather than covering the full theoretical structure space uniformly.}
    \label{fig:data_manifold}
\end{figure*}

Figure~\ref{fig:data_manifold} provides a joint manifold view of the two domains. The experimental samples overlap with the global simulated manifold, but they do not cover it uniformly. Instead, they concentrate in several localized high-density regions. The crystal-system labels and the density and volume-per-atom gradients remain spatially organized on the same manifold, which suggests that the experimental subset occupies a localized region of the full simulated structure space rather than forming a uniform subset. Taken together, the observed domain gap is reflected in symmetry composition, structural scale, chemical composition, and joint manifold coverage.

\subsection{Quantitative Decomposition}
\label{app:qualitative}

\begin{figure*}[htbp]
    \centering

    \begin{subfigure}{0.85\linewidth}
        \centering
        \includegraphics[width=1\linewidth]{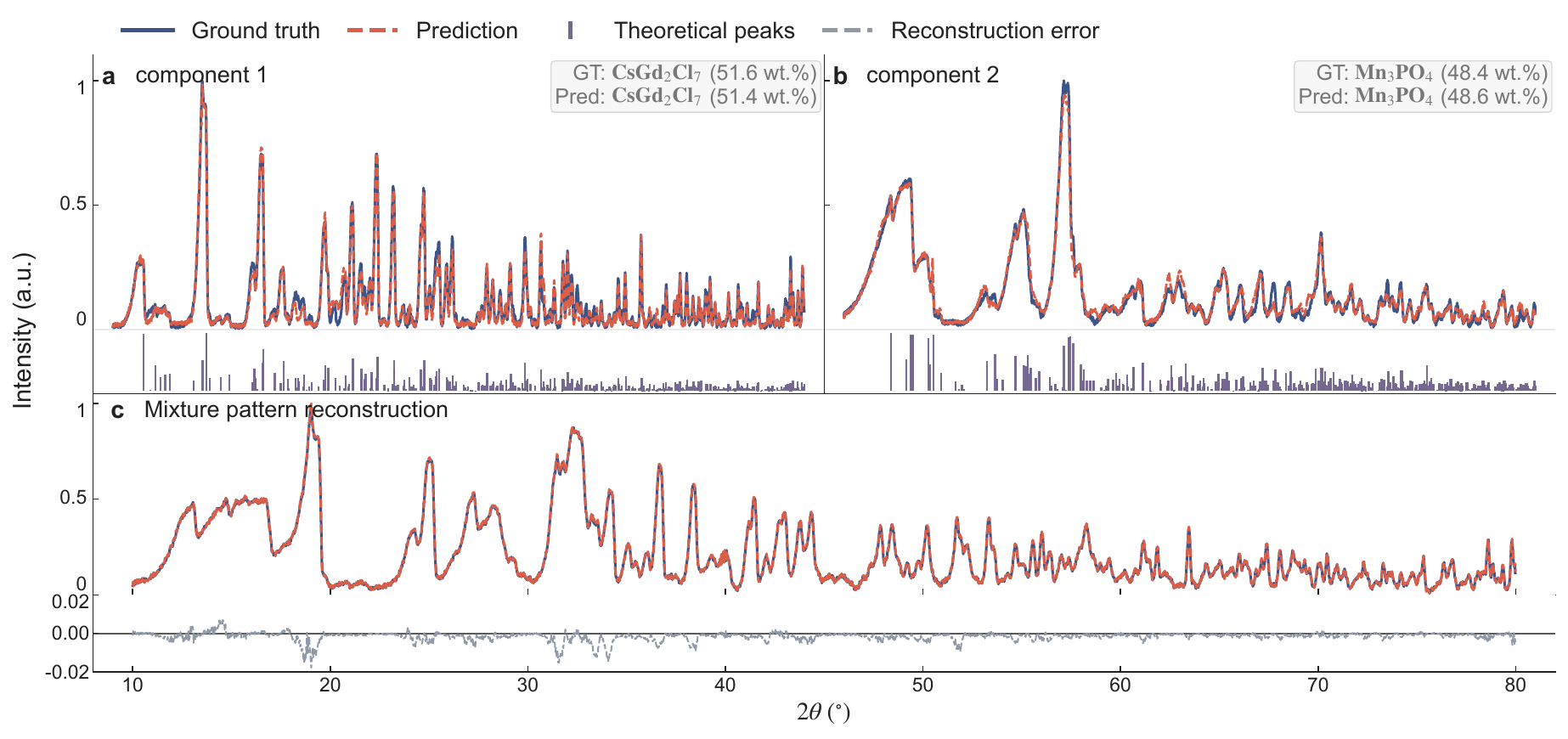}
        \caption*{\textbf{a}~Balanced mixture: CsGd$_2$Cl$_7$--MnPO$_4$}
        \label{fig:sim_k2_example1}
    \end{subfigure}

    \begin{subfigure}{0.85\linewidth}
        \centering
        \includegraphics[width=1\linewidth]{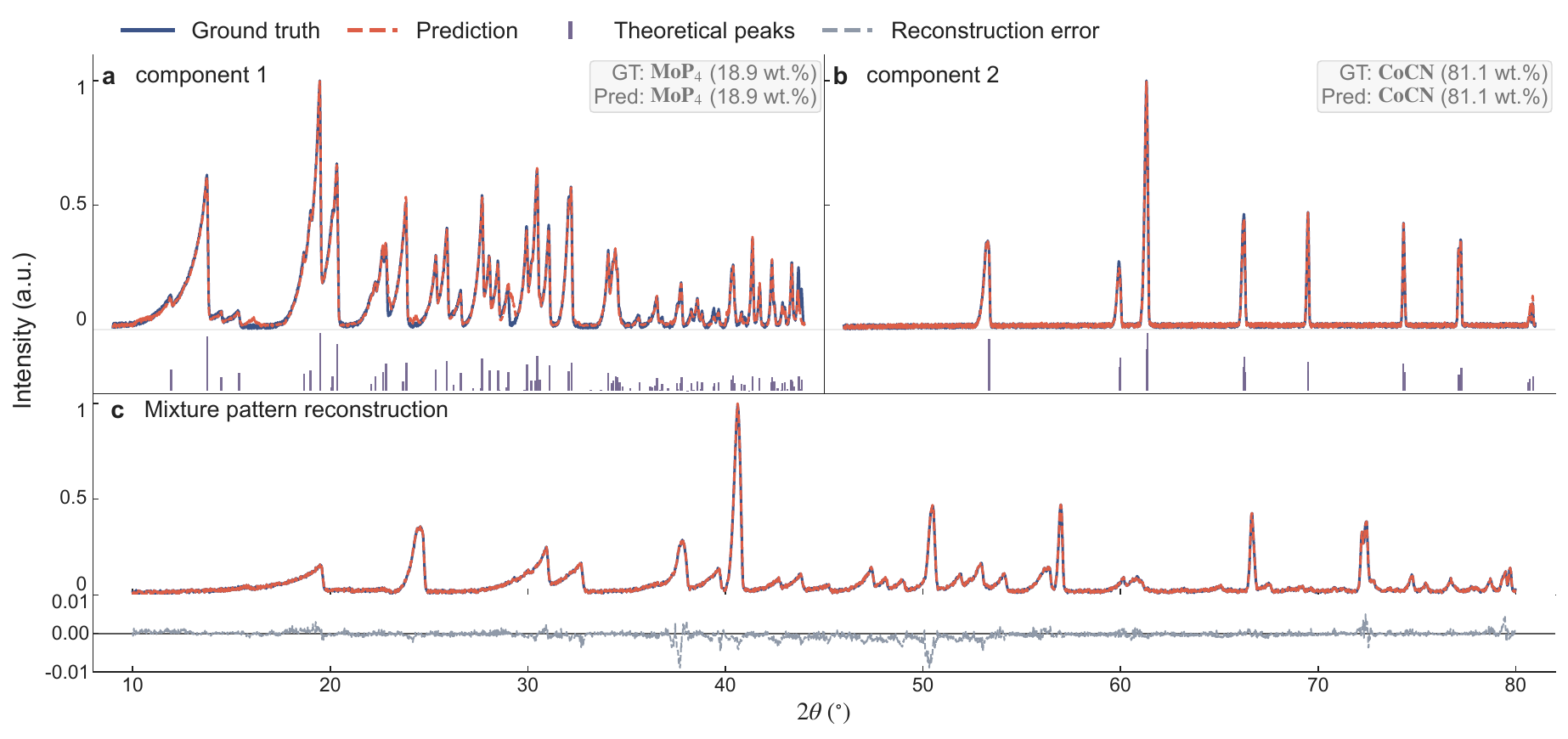}
        \caption*{\textbf{b}~Highly imbalanced mixture: MoP$_4$--CoCN}
        \label{fig:sim_k2_example2}
    \end{subfigure}

    \caption{Whole-pattern decomposition results for binary mixtures with $K=2$.}
    \label{fig:sim_k2}
\end{figure*}

\begin{figure*}[htbp]
    \centering

    \begin{subfigure}{0.85\linewidth}
        \centering
        \includegraphics[width=1\linewidth]{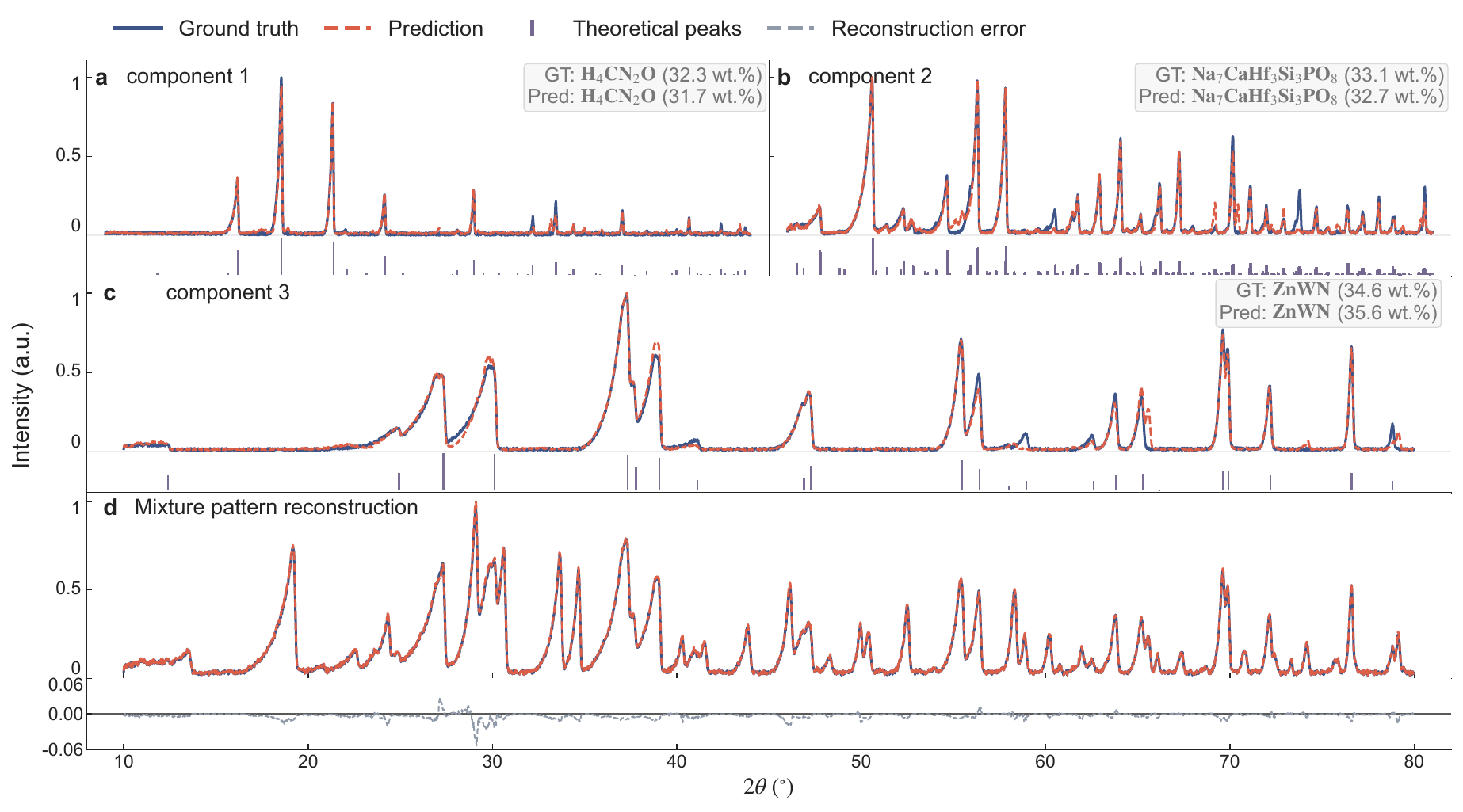}
        \caption*{\textbf{a}~Balanced mixture: H$_4$CN$_2$O--Na$_7$CaHf$_3$Si$_3$PO$_4$--ZnWN}
        \label{fig:sim_k3_example1}
    \end{subfigure}

    \begin{subfigure}{0.85\linewidth}
        \centering
        \includegraphics[width=1\linewidth]{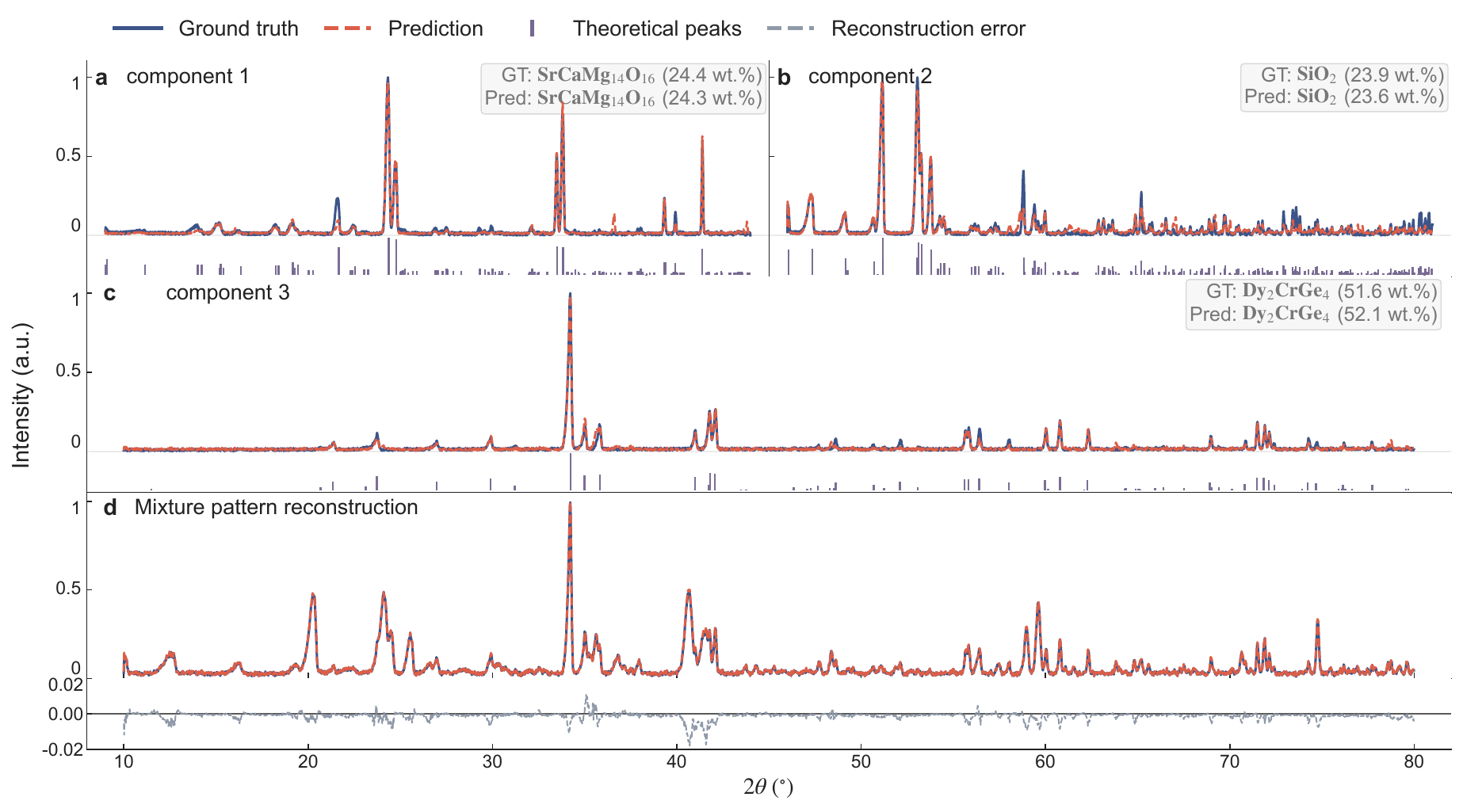}
        \caption*{\textbf{b}~Highly imbalanced mixture: SrCaMg$_{14}$O$_{16}$--SiO$_2$--Dy$_2$CrGe$_4$}
        \label{fig:sim_k3_example2}
    \end{subfigure}
    \caption{Whole-pattern decomposition results for ternary mixtures with $K=3$.}
    \label{fig:sim_k3}
\end{figure*}

\begin{figure*}[htbp]
    \centering

    \begin{subfigure}{0.85\linewidth}
        \centering
        \includegraphics[width=1\linewidth]{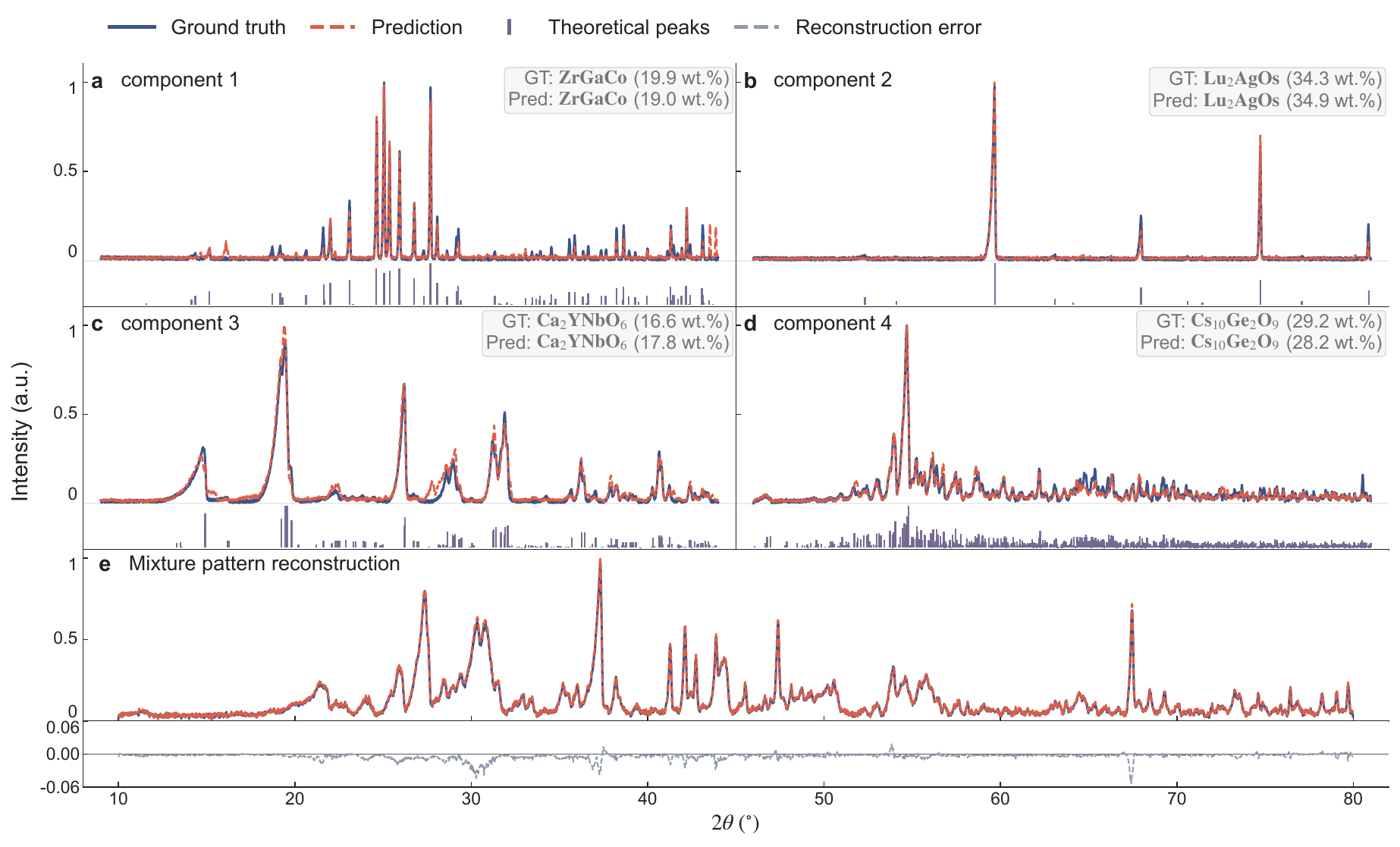}
        \caption*{\textbf{a}~ZrGaCo--Lu$_2$AgOs--Ca$_2$YNbO$_6$--Cs$_{10}$Ge$_2$O$_9$}
        \label{fig:sim_k4_example1}
    \end{subfigure}


    \begin{subfigure}{0.85\linewidth}
        \centering
        \includegraphics[width=1\linewidth]{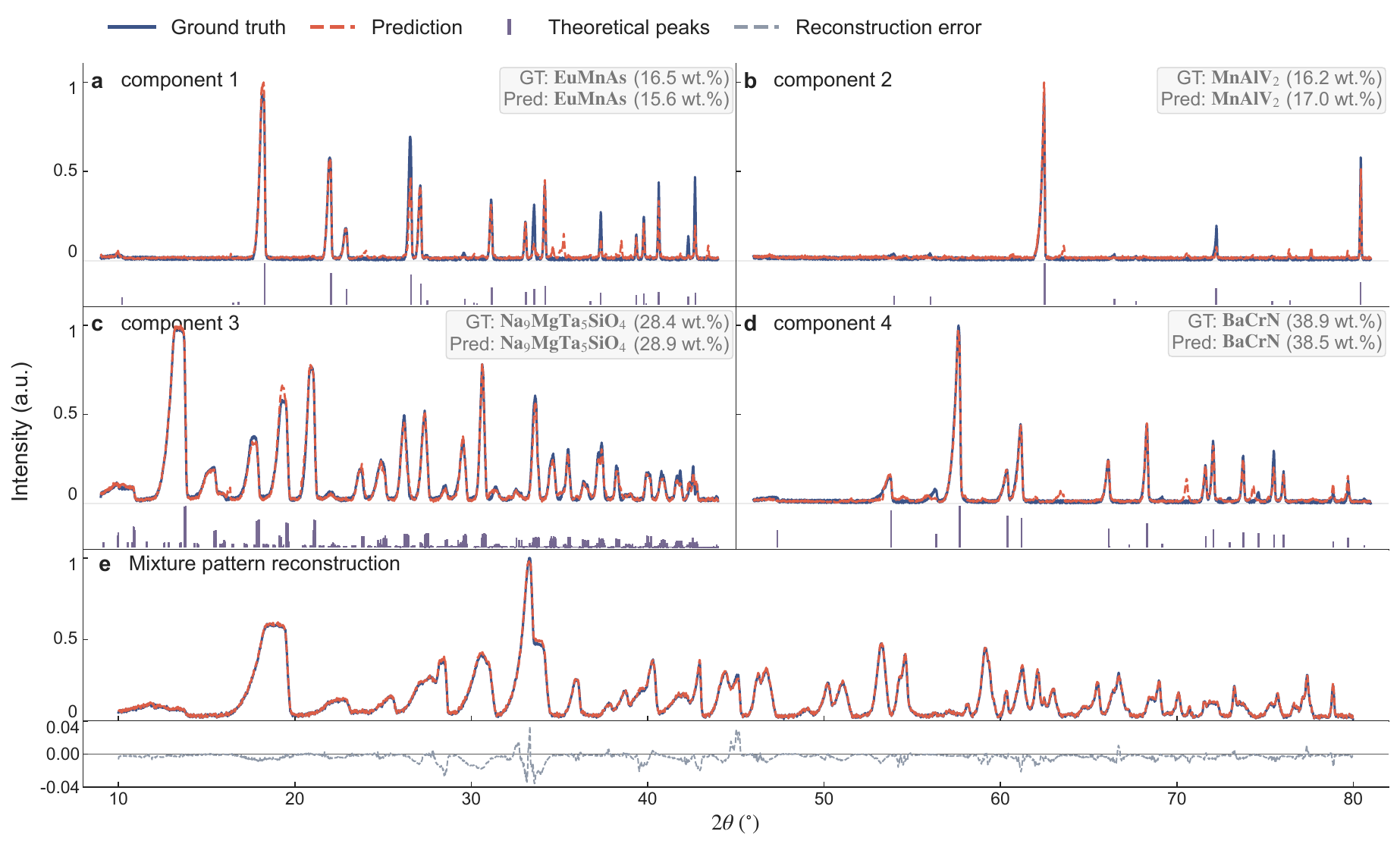}
        \caption*{\textbf{b}~EuMnAs--MnAlV$_2$--Na$_9$MgTa$_5$SiO$_4$--BaCrN}
        \label{fig:sim_k4_example2}
    \end{subfigure}

    \caption{Whole-pattern decomposition results for quaternary mixtures with $K=4$.}
    \label{fig:sim_k4}
\end{figure*}

This section presents quantitative whole-pattern decomposition results obtained by XDecomposer. Figures~\ref{fig:sim_k2}, \ref{fig:sim_k3}, and \ref{fig:sim_k4} show representative examples for mixture complexities of $K=2$, $3$, and $4$, respectively.

\section{Additional Knowledge}
\label{app:add_appendix}
\subsection{Search-Match Approach}
\label{SM_method}
The predominant method for conventional phase identification in X-ray powder diffraction patterns is the search-match approach, which involves three key steps.

First, (d, I) values, representing interplanar spacing (d) and peak intensity (I), are extracted from the diffraction pattern. Next, potential phases are identified by searching diffraction databases for matching d values, such as those in the Hanawalt indexes~\cite{UCLPXRD}. Candidate phases are then compared to the pattern’s (d, I) values using a scoring system, helping to select the best match. This process continues until satisfactory alignment is achieved for most (d, I) values. Despite advances in computer technology enabling novel qualitative methods, the core goal remains to match experimental data with database entries and compute a corresponding score.

\subsection{In-library and Out-of-library Tasks}
\label{app:In-out-tasks}

For phase identification, the ultimate goal is to map an experimental diffraction pattern, or a decomposed phase pattern in the case of multiphase materials, to a specific crystal structure. 
Depending on whether the target structure has been previously recorded, this task can be categorized into two distinct scenarios: in-library identification and out-of-library identification, as defined in our previous work \cite{bin2025simxrd}.

In-library identification aims to identify unknown materials by matching their XRD patterns against a library of known crystal structures. 
This represents the most widely used scenario in practical crystallographic analysis. Since the establishment of the Powder Diffraction File (PDF) database in 1938, experimentally determined crystal structures have been systematically collected and archived, enabling structure identification through diffraction pattern retrieval and comparison. 
With the development of computational methods, numerous search-match software packages \cite{cao2026xmatcher} have been developed to accelerate this process.
The primary challenge in this scenario is not discovering new structures, but achieving robust identification under realistic experimental variations, including instrumental noise, preferred orientation, lattice strain, peak broadening, and other measurement-related factors.

In contrast, out-of-library identification refers to the identification of crystal structures that are absent from existing diffraction databases. 
Since no reference patterns are available for direct matching, this task requires solving the inverse crystallographic problem, where diffraction patterns must be interpreted to recover an unknown three-dimensional atomic arrangement.
Typically, this involves multiple stages, including symmetry determination, estimation of lattice parameters and space groups, chemical composition inference, structure solution, and subsequent refinement and validation.
Although recent advances in generative crystal structure prediction have shown promising capabilities for exploring unknown materials, experimentally validated predictions of entirely novel crystal structures remain extremely limited.
Therefore, out-of-library identification represents a much more challenging and less frequently encountered problem compared with in-library identification.

\subsection{
Crystal Structure Determination via PXRD
}
\label{Structure-Determination}

Crystal structure determination via PXRD typically involves two main steps:
\begin{enumerate}[leftmargin=*]
    \item \textbf{Initial structure identification:} interpreting the experimental PXRD pattern to propose a plausible (partial or approximate) structure.
    \item \textbf{Structure refinement:} using the proposed structure as input for further relaxation and optimization to best fit the experimental data.
\end{enumerate}

The first step focuses on identifying a rough structural model, while the second step fine-tunes this model to capture structural details. It is well understood that no two crystals are exactly identical, even for the same compound. For example, NaCl crystals synthesized under different conditions can exhibit structural variations and defects that must be resolved through refinement.

However, refinement cannot begin from scratch. It works by minimizing the differences between the observed PXRD profile and the theoretical pattern calculated from a structural model. Once this fitting process converges, the refined parameters represent the only physically meaningful and experimentally determinable information. Therefore, an accurate initial guess of the crystal structure is essential for successful refinement. This task—\textit{structure identification}—is the focus of this paper.

The purpose of structure identification is to recognize known reference crystals from their PXRD patterns. For instance, identifying that a given pattern belongs to NaCl, without needing to determine precise lattice parameters or defects. For the past few decades, a \textbf{search-match strategy} has been widely adopted. By comparing experimental patterns against large crystal structure databases, this approach retrieves candidate structures that best match the observed data and recommends the most similar ones. This process essentially confirms the tested sample based on historical crystallographic knowledge and is referred to as \textit{in-library identification}.

Only when all known structures fail to match does \textit{out-of-library identification} begin. Traditionally, this is a labor-intensive process involving the determination of lattice geometry, ideal chemical composition, and iterative refinement with trial and error. In recent years, generative methods, those that propose structures from scratch, have been increasingly applied in such situations. However, experimental crystal structure determination remains highly challenging, and generation from scratch carries a high risk of producing incorrect structures.

Current methods mainly focus on single-phase identification, where a PXRD pattern contains only one dominant phase. However, for the multiphase scenario considered in this work, most existing approaches still rely on an iterative subtract-and-match strategy: a candidate phase is first identified from the observed mixture, then its contribution is estimated and subtracted, and the residual pattern is used for the next round of matching and identification.
This strategy is essentially a trial-and-error process and is highly sensitive to errors in early-stage phase identification. In multiphase mixtures, strong peak overlap and mutual interference among components greatly reduce the reliability of prior phase identification. Even when a candidate phase is correctly retrieved, its subtraction is often inaccurate because the true phase proportion is unknown, and the theoretical reference pattern may differ significantly from its actual experimental contribution due to peak broadening, background effects, preferred orientation, and instrumental variation.

As a result, error accumulation is common, and the residual pattern may become increasingly unreliable for subsequent identification steps. This limitation highlights that multiphase identification should not be treated as a recursive single-phase matching problem. Instead, it requires joint decomposition of all constituent single-phase signals directly from the mixed pattern, enabling simultaneous recovery of component contributions and more reliable downstream phase identification.

\clearpage
\end{document}